\documentclass{article}
\usepackage{microtype}
\usepackage{graphicx}
\usepackage{subfigure}
\usepackage{booktabs}
\usepackage{amsmath}
\usepackage{amsthm}
\usepackage{amssymb}
\usepackage{colortbl}
\usepackage{amsfonts}
\usepackage{hyperref}
\usepackage[accepted]{icml2021}

\newtheorem{thm}{Theorem}

\newtheorem*{thm*}{Theorem}
\newtheorem{cor}{Corollary}

\newtheorem{lem}{Lemma}

\theoremstyle{remark}
\newtheorem{rem}{Remark}

\icmltitlerunning{Adversarial Robustness Guarantees for Random Deep Neural Networks}

\begin{document}

\twocolumn[
\icmltitle{Adversarial Robustness Guarantees for Random Deep Neural Networks}

\begin{icmlauthorlist}
\icmlauthor{Giacomo De Palma}{SNS,MechE,RLE}
\icmlauthor{Bobak T. Kiani}{RLE,EECS}
\icmlauthor{Seth Lloyd}{MechE,RLE}
\end{icmlauthorlist}

\icmlaffiliation{SNS}{Scuola Normale Superiore, Pisa, Italy}
\icmlaffiliation{MechE}{Department of Mechanical Engineering, MIT, Cambridge MA, USA}
\icmlaffiliation{RLE}{Research Laboratory of Electronics, MIT, Cambridge MA, USA}
\icmlaffiliation{EECS}{Department of Electrical Engineering \& Computer Science, MIT, Cambridge MA, USA}

\icmlcorrespondingauthor{Giacomo De Palma}{giacomo.depalma@sns.it}
\icmlcorrespondingauthor{Bobak T. Kiani}{bkiani@mit.edu}
\icmlcorrespondingauthor{Seth Lloyd}{slloyd@mit.edu}

\icmlkeywords{Deep neural networks, adversarial examples, Gaussian processes}

\vskip 0.3in
]

\printAffiliationsAndNotice{}

\begin{abstract}
The reliability of deep learning algorithms is fundamentally challenged by the existence of adversarial examples, which are incorrectly classified inputs that are extremely close to a correctly classified input.
We explore the properties of adversarial examples for deep neural networks with random weights and biases, and prove that for any $p\ge1$, the $\ell^p$ distance of any given input from the classification boundary scales as one over the square root of the dimension of the input times the $\ell^p$ norm of the input.
The results are based on the recently proved equivalence between Gaussian processes and deep neural networks in the limit of infinite width of the hidden layers, and are validated with experiments on both random deep neural networks and deep neural networks trained on the MNIST and CIFAR10 datasets.
The results constitute a fundamental advance in the theoretical understanding of adversarial examples, and open the way to a thorough theoretical characterization of the relation between network architecture and robustness to adversarial perturbations.
\end{abstract}

\section{Introduction}\label{sec:intro}
Deep neural networks constitute an extremely powerful architecture for machine learning and have achieved an enormous success in several fields such as speech recognition, computer vision and natural language processing where they can often outperform human abilities \cite{mnih2015human,lecun2015deep,radford2015unsupervised,schmidhuber2015deep,goodfellow2016deep}.
In 2014, a very surprising property of deep neural networks emerged in the context of image classification \cite{szegedy2013intriguing,goodfellow2014explaining}: an extremely small perturbation can change the label of a correctly classified image.
This property poses serious challenges to the reliability of deep learning algorithms since it may be exploited by a malicious adversary to fool a machine learning algorithm by steering its output.
For this reason, methods to find perturbed inputs or adversarial examples have been named adversarial attacks.
This problem further captured the attention of the deep learning community when it was discovered that real-world images taken with a camera can also constitute adversarial examples \cite{kurakin2018adversarial,sharif2016accessorize,brown2017adversarial,evtimov2017robust}.
To study adversarial attacks, two lines of research have been developed: one aims at developing efficient algorithms to find adversarial examples \cite{su2019one,athalye2018synthesizing,liu2016delving}, and the other aims at making deep neural networks more robust against adversarial attacks \cite{madry2018towards,tsipras2018robustness,nakkiran2019adversarial,lecuyer2019certified,gilmer2019adversarial}; algorithms to compute the robustness of a given trained deep neural network against adversarial attacks have also been developed \cite{li2019certified,jordan2019provable}.

Several theories have been proposed to explain the phenomenon of adversarial examples
\cite{raghunathan2018certified,wong2018provable,xiao2018training,cohen2019certified,schmidt2018adversarially,tanay2016boundary,kim2019bridging,fawzi2016robustness,shamir2019simple,bubeck2019adversarial,ilyas2019adversarial}.
One of the most prominent theories states that adversarial examples are an unavoidable feature of the high-dimensional geometry of the input space: Refs. \cite{gilmer2018adversarial,fawzi2018adversarial,shafahi2019adversarial,mahloujifar2019curse} show that, whenever the classification error is finite, the label of a correctly classified input can be changed with an adversarial perturbation of size $O\left(1\left/\sqrt{n}\right.\right)$ times the norm of the input, where $n$ is the dimension of the input space.

In this paper, we explore the properties of adversarial examples for deep neural networks with random weights and biases in the limit of infinite width of the hidden layers.
Our main result, presented in \autoref{sec:main}, is a probabilistic robustness guarantee on the $\ell^1$ distance of a given input from the closest classification boundary, which we later extend to all the $\ell^p$ distances\footnote{the $\ell^p$ norm of a vector $x\in\mathbb{R}^n$ is $\left\|x\right\|_p = \left(\sum_{i=1}^n\left|x_i\right|^p\right)^\frac{1}{p}$.}.
We prove that the $\ell^1$ distance from the closest classification boundary of any given input $x\in\mathbb{R}^n$ whose entries are $O(1)$ is with high probability at least $\tilde{\Omega}\left(\sqrt{n}\right)$ (the tilde means that logarithmic factors are hidden), \emph{i.e.}, the distance of any adversarial example from $x$ is larger than $\tilde{\Omega}\left(\sqrt{n}\right)$.
Since $\|x\|_1 = \Theta(n)$, our result implies that the size of any adversarial perturbation is at least $\tilde{\Omega}\left(1\left/\sqrt{n}\right.\right)$ times the norm of the input.
This lower bound to the size of adversarial perturbations matches the upper bound imposed by the high-dimensional geometry proven in Refs. \cite{gilmer2018adversarial,fawzi2018adversarial,shafahi2019adversarial,mahloujifar2019curse}.
Therefore, our result proves that $1\left/\sqrt{n}\right.$ is the universal scaling of the minimum size of adversarial perturbations with respect to the norm of the input.
We also prove that, for any given unit vector $v\in\mathbb{R}^n$, with high probability all the inputs $x+t\,v$ with $0\le t\le O\left(\sqrt{n}\right)$ have the same classification as $x$.
Since $\|x\|_2 = \Theta\left(\sqrt{n}\right)$, a remarkable consequence of this result is that a finite fraction of the $\ell^2$ distance to the origin can be traveled without encountering any classification boundary.

Our results encompass a wide variety of network architectures, namely any combination of convolutional or fully connected layers with nonlinear activation, skipped connections and pooling (see \autoref{sec:setup}).
Our proof builds on the recently proved equivalence between deep neural networks with random weights and biases in the limit of infinite width and Gaussian processes \cite{lee2018deep,yang2019wide}.
We prove that the same probabilistic robustness guarantees for the adversarial distance also apply to a broad class of Gaussian processes when the variance is lower bounded by the Euclidean square norm of the input and the feature map of the kernel associated to the Gaussian process has an $O(1)$ Lipschitz constant, a result that can be of independent interest.

In \autoref{sec:adv_attack_random}, we experimentally validate our theoretically predicted scaling of the adversarial distance for random deep neural networks, and we find a very good agreement between theory and experiments starting from $n\gtrsim100$.
In \autoref{sec:adv_attack_trained}, we perform experiments on the adversarial distance for deep neural networks trained on the MNIST and CIFAR10 datasets.
In both cases, the training does not change the order of magnitude of the adversarial distance.
While for MNIST the adversarial distances for random and trained networks are very close, in the case of CIFAR10 the training decreases the adversarial distance by roughly half order of magnitude.
As better discussed in \autoref{sec:adv_attack_trained}, this can be ascribed to the different nature of the CIFAR10 with respect to the MNIST data.

Our adversarial robustness guarantee applies also to deep neural networks trained with Bayesian inference under the hypothesis that the target function $f$ is generated by the same random deep neural network employed for the training.
Indeed, given the training inputs $x^{(1)},\,\ldots,\, x^{(n)}$ (which do not need to be random) and the corresponding random training labels $y^{(1)}=f\left(x^{(1)}\right),\,\ldots,\,y^{(n)}=f\left(x^{(n)}\right)$, the Bayesian classifier is a function $g$ randomly drawn from the posterior probability distribution $q$ obtained by conditioning on the observation of $y^{(1)},\,\ldots,\, y^{(n)}$ the prior probability distribution $p$ generated by the random deep neural network. If we forget the values of the training labels, the probability distribution of $g$ becomes the average of $q$ over the possible values of the training labels $y^{(1)},\,\ldots,\, y^{(n)}$ induced by the random target function $f$.
Such average has the effect of removing the conditioning on $y^{(1)},\,\ldots,\, y^{(n)}$. Therefore, after such average, the probability distribution of the Bayesian classifier $g$ coincides with the prior probability distribution $p$ regardless of the choice of $p$ and of $x^{(1)},\,\ldots,\, x^{(n)}$.
Therefore, the properties of the adversarial distance for a given random deep neural network and for the Bayesian classifier associated to the same network coincide under the hypothesis that the target function is also generated by the same random network.

\subsection{Related Works}
The equivalence between Gaussian processes and neural networks with random weights and biases in the limit of infinite width of the hidden layers has been known for a long time in the case of fully connected neural networks with one hidden layer \cite{neal1996priors,williams1997computing}, and has recently been extended to multi-layer \cite{schoenholz2016deep,pennington2018emergence,lee2018deep,matthews2018gaussian,poole2016exponential,schoenholz2016deep} and convolutional deep neural networks \cite{garriga-alonso2018deep,xiao2018dynamical,novak2019bayesian}.
The equivalence is now proved for practically all the existing neural networks architectures \cite{yang2019wide}, and has been extended to trained deep neural networks \cite{jacot2018neural,lee2019wide,yang2019scaling,arora2019exact,huang2019dynamics,li2019enhanced,wei2019regularization,cao2019generalization} including adversarial training \cite{gao2019convergence}.
Ref. \cite{cardelli2019robustness} proves a probabilistic robustness guarantee for Bayesian classifiers with prior probability distribution given by a Gaussian process in the same spirit of \autoref{thm:main}, and also this proof exploits the Borell--TIS inequality and Dudley's theorem.
The results of \cite{cardelli2019robustness} have been expanded to probabilistic guarantees for neural networks in \cite{cardelli2019statistical,wicker2020probabilistic}.
The smoothness of the feature map of a kernel plays a key role in machine learning applications \cite{mallat2012group,oyallon2015deep,bruna2013invariant,bietti2019group} and kernels associated to deep neural networks have been studied from this point of view \cite{bietti2019inductive}.
In the setup of binary classification of bit strings, the Hamming distance of a given input from the closest classification boundary has been theoretically studied in \cite{de2019random}, where the scaling $O\left(\sqrt{n\left/\ln n\right.}\right)$ has been found.

\section{Setup}\label{sec:setup}
Our inputs are $D$-dimensional images considered as elements of $\mathbb{R}^{n_C^{(0)}\times\mathcal{I}^{(0)}}$, where $n_C^{(0)}$ is the number of the input channels (e.g., $n_C^{(0)}=3$ for Red-Green-Blue images) and $\mathcal{I}^{(0)} = \mathbb{Z}_{h_1}\times\ldots\times\mathbb{Z}_{h_D}$ is the set of the input pixels, assumed for simplicity to be periodic.
$D=2$ recovers standard 2D images.
For the sake of a simpler notation, we will sometimes consider the input space as $\mathbb{R}^n$, with $n = n_C^{(0)}\left|\mathcal{I}^{(0)}\right|$.

Our architecture allows for any combination of convolutional layers, fully connected layers, skipped connections and pooling.
For the sake of a simpler notation, we treat each of the above operations as a layer, even if it does not include any nonlinear activation.
For simplicity, we assume that the nonlinear activation function is the ReLU $\tau(x) = \max(0,x)$.
Our results can be easily extended to other activation functions.

For any $l=1,\,\ldots,\,L+1$ and any input $x\in\mathbb{R}^{n_C^{(0)}\times\mathcal{I}^{(0)}}$, let $n_C^{(l)}$ be the number of channels and $\mathcal{I}^{(l)}$ the set of pixels of the output of the $l$-th layer $\phi^{(l)}(x)\in\mathbb{R}^{n_C^{(l)}\times\mathcal{I}^{(l)}}$.
The layer transformations have the following mathematical expression:
\begin{itemize}
\item {\bf Input layer:}
We have $\mathcal{I}^{(1)} = \mathcal{I}^{(0)}$ and
\begin{equation}\label{eq:phi1}
\phi^{(1)}_{i,\alpha}(x) = b_i^{(1)} + \sum_{j=1}^{n_C^{(0)}}\sum_{\beta\in\mathcal{P}^{(1)}}W_{ij,\beta}^{(1)}\,x_{j,\alpha+\beta}
\end{equation}
for any $i=1,\,\ldots,\,n_C^{(1)}$ and any $\alpha\in\mathcal{I}^{(1)}$, where $\mathcal{P}^{(1)}\subseteq\mathcal{I}^{(1)} = \mathcal{I}^{(0)}$ is the convolutional patch of the first layer.
We assume for simplicity that $-\mathcal{P}^{(1)} = \mathcal{P}^{(1)}$.
\item {\bf Nonlinear layer:}
If the ($l+1$)-th layer is a nonlinear layer, we have $\mathcal{I}^{(l+1)} = \mathcal{I}^{(l)}$ and
\begin{align}\label{eq:phil+1}
&\phi^{(l+1)}_{i,\alpha}(x) =\nonumber\\
&b^{(l+1)}_i + \sum_{j=1}^{n_C^{(l)}}\sum_{\beta\in\mathcal{P}^{(l+1)}}W_{ij,\beta}^{(l+1)}\,\tau\left(\phi^{(l)}_{j,\alpha-\beta}(x)\right)
\end{align}
for any $i=1,\,\ldots,\,n_C^{(l+1)}$ and any $\alpha\in\mathcal{I}^{(l+1)}$, where $\tau:\mathbb{R}\to\mathbb{R}$ is the activation function and $\mathcal{P}^{(l+1)}\subseteq\mathcal{I}^{(l+1)} = \mathcal{I}^{(l)}$ is the convolutional patch of the layer.
We assume for simplicity that $-\mathcal{P}^{(l+1)}=\mathcal{P}^{(l+1)}$.
Fully connected layers are recovered by $\left|\mathcal{I}^{(l)}\right| = \left|\mathcal{I}^{(l+1)}\right| = \left|\mathcal{P}^{(l+1)}\right| = 1$.
\item {\bf Skipped connection:}
If the ($l+1$)-th layer is a skipped connection, we have $n_C^{(l+1)}=n_C^{(l)}$, $\mathcal{I}^{(l+1)} = \mathcal{I}^{(l)}$ and
\begin{equation}\label{eq:skip}
\phi^{(l+1)}_{i,\alpha}(x) = \phi^{(l)}_{i,\alpha}(x) + \phi^{(l-k)}_{i,\alpha}(x)
\end{equation}
for any $i=1,\,\ldots,\,n_C^{(l+1)}$ and any $\alpha\in\mathcal{I}^{(l+1)}$, where $k\in\left\{1,\,\ldots,\,l-2\right\}$ is such that the sum in \eqref{eq:skip} is well defined, i.e., $n^{(l-k)}_C = n^{(l)}_C$ and $\mathcal{I}^{(l-k)} = \mathcal{I}^{(l)}$.
For the sake of a simple proof, we assume that the $l$-th layer is either a convolutional or a fully connected layer.
\item {\bf Pooling:}
If the ($l+1$)-th layer is a pooling layer, we have $n_C^{(l+1)} = n_C^{(l)}$, and $\mathcal{I}^{(l+1)}$ is a partition of $\mathcal{I}^{(l)}$, i.e., the elements of $\mathcal{I}^{(l+1)}$ are disjoint subsets of $\mathcal{I}^{(l)}$ whose union is equal to $\mathcal{I}^{(l)}$.
We assume for simplicity that the $l$-th layer is a convolutional layer and that all the elements of $\mathcal{I}^{(l+1)}$ have the same cardinality, which is therefore equal to $\left.\left|\mathcal{I}^{(l)}\right|\right/\left|\mathcal{I}^{(l+1)}\right|$.
We have
\begin{equation}
\phi^{(l+1)}_{i,\alpha}(x) = \sum_{\beta\in\alpha}\phi^{(l)}_{i,\beta}(x)
\end{equation}
for any $i=1,\,\ldots,\,n_C^{(l+1)}$ and any $\alpha\in\mathcal{I}^{(l+1)}$.
\item {\bf Flattening layer:}
Let the ($L_f+1$)-th layer be the flattening layer. We notice that we include a fully connected layer directly after the flattening as part of this layer.
We have $\left|\mathcal{I}^{(L_f+1)}\right|=1$ and
\begin{align}
&\phi^{(L_f+1)}_i(x) = \nonumber\\
&b_i + \sum_{j=1}^{n_C^{(L_f)}}\sum_{\alpha\in\mathcal{I}^{(L_f)}}W^{(L_f+1)}_{ij,\alpha}\,\tau\left(\phi^{(L_f)}(x)\right)
\end{align}
for any $i=1,\,\ldots,\,n_C^{(L_f+1)}$.
\item {\bf Output layer:}
The final output of the network is $\phi(x) = \phi^{(L+1)}_1(x)$, and the output label is $\mathrm{sign}\,\phi(x)$.
We introduce the other components of $\phi^{(L+1)}$ for the sake of a simpler notation in the proof of \autoref{thm:DNN}.
\end{itemize}
Our random deep neural networks draw all the weights $W^{(l)}_{ij,\alpha}$ and the biases $b^{(l)}_i$ from independent Gaussian probability distributions with zero mean and variances $\left.{\sigma^{(l)}_W}^2\right/n_C^{(l-1)}$ and ${\sigma^{(l)}_b}^2$, respectively.
The variances are allowed to depend on the layer.

\section{Theoretical Results}\label{sec:main}
\label{sec:main_results}
A recent series of works \cite{schoenholz2016deep,pennington2018emergence,lee2018deep,matthews2018gaussian,poole2016exponential, garriga-alonso2018deep,xiao2018dynamical,novak2019bayesian,yang2019wide} has proved that in the limit $n_C^{(1)},\,\ldots,\,n_C^{(L+1)}\to\infty$ the random deep neural networks defined in \autoref{sec:setup} are centered Gaussian processes, i.e., for any $M\in\mathbb{N}$ and any set of $M$ inputs $x^1,\,\ldots,\,x^M\in\mathbb{R}^{n_C^{(0)}\times\mathcal{I}^{(0)}}$, the joint probability distribution of the corresponding outputs $\phi\left(x^1\right),\,\ldots,\,\phi\left(x^M\right)\in\mathbb{R}$ is Gaussian with zero mean and covariance given by a kernel $K(x,y) = \mathbb{E}(\phi(x)\,\phi(y))$ that depends on the architecture of the deep neural network.
Therefore, the properties of adversarial perturbations for random deep neural networks are equivalent to the properties of adversarial perturbations for the corresponding Gaussian processes.
First, we prove in \autoref{thm:main} an adversarial robustness guarantee for a broad class of Gaussian processes.
We then prove in \autoref{thm:DNN} that the guarantee applies to the Gaussian processes generated by random deep neural networks, and therefore it applies to random deep neural networks.

We recall that we can associate to any kernel $K$ on $\mathbb{R}^n$ a Reproducing Kernel Hilbert Space (RKHS) $\mathcal{H}$ with scalar product and norm denoted by $\cdot$ and $\left\|\cdot\right\|$, respectively, and a feature map $\Phi:\mathbb{R}^n\to\mathcal{H}$ such that for any $x,\,y\in\mathbb{R}^n$ \cite{rasmussen2006gaussian}
\begin{equation}
K(x,y) = \Phi(x)\cdot \Phi(y)\,.
\end{equation}
The kernel $K$ induces on the input space the RKHS distance
\begin{align}
{d(x,y)}^2 &= \left\|\Phi(x)-\Phi(y)\right\|^2\nonumber\\
&= K(x,x) -2\,K(x,y) + K(y,y)\,.
\end{align}

We can now state our main result.
\begin{thm}[$\ell^1$ adversarial robustness guarantee for Gaussian processes]\label{thm:main}
Let $\phi$ be a Gaussian process on $\mathbb{R}^n$ with zero mean and covariance $K$, and let $d$ be the associated RKHS distance.
Let $C,\,M>0$ be such that for any $x,\,y\in\mathbb{R}^n$
\begin{equation}\label{eq:C12}
\sqrt{K(x,x)} \ge C\left\|x\right\|_2\,,\quad d(x,y) \le M\,C\left\|x-y\right\|_2\,.
\end{equation}
Let $x_0\in\mathbb{R}^n$, and for any $r>0$ let
\begin{equation}
\mathcal{B}_r^1 = \left\{x\in \mathbb{R}^n:\|x-x_0\|_1<r\right\}
\end{equation}
be the $\ell^1$ ball with center $x_0$ and radius $r$.
Then,
for any $0<\delta<1$ and any
\begin{equation}\label{eq:r}
0<r\le \frac{\left\|x_0\right\|_2\delta\sqrt{\pi}}{M\left(12\sqrt{\ln 4n} + 8\ln n\,\sqrt{\ln 2n} + 2\sqrt{\pi}\right)}
\end{equation}
we have
\begin{equation}
\mathbb{P}\left(\exists\, x\in \mathcal{B}_r^1:\phi(x)=0\right) \le \delta\,.
\end{equation}
Moreover, let $v$ be a unit vector in $\mathbb{R}^n$, and for any $r>0$ let $\mathcal{L}_r=\left\{x_0 + t\,v:0\le t \le r\right\}$ be the segment starting in $x_0$, parallel to $v$ and with length $r$.
Then, for any
\begin{equation}\label{eq:r2}
0 < r \le \pi\left\|x_0\right\|_2\delta\left/\left(2\,M + \pi\right)\right.
\end{equation}
we have
\begin{equation}
\mathbb{P}\left(\exists\, x\in \mathcal{L}_r : \phi(x)=0\right)\le\delta\,.
\end{equation}
\end{thm}
We prove the first part of \autoref{thm:main} in \autoref{sec:mainproofI}, and we refer to \autoref{sec:mainproofII} for the proof of the second part.
\begin{rem}
Since our classifier is $\mathrm{sign}\,\phi(x)$, we have $\phi(x)=0$ for some $x$ in $\mathcal{B}_r^1$ iff $\mathcal{B}_r^1$ is crossed by a classification boundary, \emph{i.e.}, iff there exists $x\in \mathcal{B}_r^1$ such that $\phi(x)\,\phi(x_0)<0$.
\end{rem}
\begin{rem}
The prefactor in \eqref{eq:r} is not sharp.
Indeed, the proof of \autoref{thm:main} relies on Dudley's theorem \cite{bartlett2013theoretical}, which provides an upper bound to the expectation value of the maximum of a Gaussian process over a given region, and on an estimate of the covering number of the $\ell^1$ unit ball (\autoref{thm:Nepsilon}). Despite employing the best state-of-the-art tools, the prefactors of both these results are not sharp \cite{ledoux2013probability,price2016sublinear}.
\end{rem}
The following \autoref{thm:DNN}, which we prove in \autoref{sec:DNNproof}, states that the kernels of the Gaussian processes associated to random deep neural networks satisfy the hypotheses of \autoref{thm:main}.
\begin{thm}[smoothness of the DNN Gaussian processes]\label{thm:DNN}
The kernel associated to the output of a random deep neural network as in \autoref{sec:setup} satisfies \eqref{eq:C12} with
\begin{equation}
M = \sqrt{\left.\left|\mathcal{I}^{(0)}\right|\right/\left|\mathcal{I}^{(L_f)}\right|}\,,
\end{equation}
where $\mathcal{I}^{(0)}$ and $\mathcal{I}^{(L_f)}$ are sets of the pixels of the input and of the layer immediately before the flattening, respectively.
\end{thm}
\begin{cor}[$\ell^1$ adversarial robustness guarantee for random deep neural networks]\label{cor:main}
Let $\phi$ be a random deep neural network as in \autoref{sec:setup}.
For any input $x_0\in\mathbb{R}^{n_C^{(0)}\times\mathcal{I}^{(0)}}$ and any $r>0$ let
\begin{equation}
\mathcal{B}_r^1 = \left\{x\in \mathbb{R}^{n_C^{(0)}\times\mathcal{I}^{(0)}}:\|x-x_0\|_1<r\right\}\,.
\end{equation}
Then, in the limit $n_C^{(1)},\,\ldots,\,n_C^{(L+1)}\to\infty$,
for any $0<\delta<1$ and any
\begin{equation}\label{eq:rn}
0<r\le \frac{\left\|x_0\right\|_2\delta\sqrt{\pi\left.\left|\mathcal{I}^{(L_f)}\right|\right/\left|\mathcal{I}^{(0)}\right|}}{12\sqrt{\ln 4n} + 8\ln n\,\sqrt{\ln 2n} + 2\sqrt{\pi}}\,,
\end{equation}
where $n = n_C^{(0)}\left|\mathcal{I}^{(0)}\right|$, we have
\begin{equation}
\mathbb{P}\left(\exists\, x\in \mathcal{B}_r^1:\phi(x)=0\right) \le \delta\,.
\end{equation}
Moreover, let $v$ be a unit vector in $\mathbb{R}^{n_C^{(0)}\times\mathcal{I}^{(0)}}$, and for any $r>0$ let $\mathcal{L}_r=\left\{x_0 + t\,v:0\le t \le r\right\}$.
Then, for any
\begin{equation}\label{eq:rn2}
0<r\le \frac{\pi\left\|x_0\right\|_2\delta}{2\sqrt{\left.\left|\mathcal{I}^{(0)}\right|\right/\left|\mathcal{I}^{(L_f)}\right|} + \pi}
\end{equation}
we have
\begin{equation}
\mathbb{P}\left\{\exists\, x\in \mathcal{L}_r : \phi(x)=0\right\}\le\delta\,.
\end{equation}
\end{cor}
\begin{rem}
The bounds of \autoref{cor:main} do not depend on the choice of the variances of weights and biases.
\end{rem}
\begin{rem}[asymptotic scaling]\label{rem:tilde}
\autoref{thm:main} and \autoref{cor:main} hold for any choice of $n$, $n_C^{(0)}$ and $\mathcal{I}^{(0)}$.
In the limit $n\to\infty$, if all the entries of $x_0$ are $\Theta(1)$ we have $\|x_0\|_2 = \Theta\left(\sqrt{n}\right)$, and therefore both \eqref{eq:r} and \eqref{eq:r2} become, up to logarithmic factors,
\begin{equation}
0<r\le \tilde{O}\left(\left.\delta\sqrt{n}\right/M\right)\,.
\end{equation}
Analogously, in the limit $n = n_C^{(0)}\left|\mathcal{I}^{(0)}\right|\to\infty$ both \eqref{eq:rn} and \eqref{eq:rn2} become
\begin{equation}
0<r\le \tilde{O}\left(\delta\sqrt{n\left.\left|\mathcal{I}^{(L_f)}\right|\right/\left|\mathcal{I}^{(0)}\right|}\right)\,.
\end{equation}
\end{rem}
\begin{rem}[$\ell^p$ adversarial robustness guarantees] \label{rem:norm_p_scaling}
Let us assume for simplicity that $\left.\left|\mathcal{I}^{(L_f)}\right|\right/\left|\mathcal{I}^{(0)}\right|$ does not scale with $n$.
For any $p\ge1$ and any $r>0$, let
\begin{equation}
\mathcal{B}_r^p = \left\{x\in \mathbb{R}^{n_C^{(0)}\times\mathcal{I}^{(0)}}:\|x-x_0\|_p<r\right\}
\end{equation}
be the $\ell^p$ ball with center $x_0$ and radius $r$.
Since
\begin{equation}
\|x-x_0\|_1 \le n^\frac{p-1}{p}\|x-x_0\|_p
\end{equation}
for any $x\in\mathbb{R}^{n_C^{(0)}\times\mathcal{I}^{(0)}}$, we trivially have from \autoref{rem:tilde} that in the limit $n\to\infty$, \begin{equation}
\mathbb{P}\left(\exists\, x\in \mathcal{B}_r^p:\phi(x)=0\right) \le \delta
\end{equation}
for $0<r\le \tilde{O}\left(\delta\,n^{\frac{1}{p} - \frac{1}{2}}\right)$.
In particular, the $\ell^2$ and $\ell^\infty$ distances from the closest classification boundary scale at least as $\tilde{\Omega}(1)$ and $\tilde{\Omega}\left(1/\sqrt{n}\right)$, respectively.
If all the entries of $x_0$ are $\Theta(1)$, then $\|x_0\|_p = \Theta(n^\frac{1}{p})$, and the ratio between the $\ell^p$ distance to the classification boundary and $\|x_0\|_p$ scales at least as $\tilde{\Omega}\left(1/\sqrt{n}\right)$ regardless of $p$, \emph{i.e.}, regardless of the choice of the norm, a fraction $1/\sqrt{n}$ of the input must be changed to change the label.
\end{rem}
To summarize, we have proved that the $\ell^1$ distance of any given input from the closest classification boundary is with high probability at least $\Omega(\sqrt{n})$, where $n$ is the dimension of the input.
Moreover, for any $p\ge1$, the $\ell^p$ distance of any given input from the closest classification boundary is with high probability at least $\Omega\left(1/\sqrt{n}\right)$ times the $\ell^p$ norm of the input.
This result applies to both smooth Gaussian processes and deep neural networks with almost any architecture and random weights and biases.

\subsection{Proof of \autoref{thm:main}, Part I}\label{sec:mainproofI}
Let
\begin{equation}
p_r=\mathbb{P}\left(\exists\, x\in \mathcal{B}_r^1:\phi(x)=0\right)\,,
\end{equation}
and for any $\phi_0>0$ let
\begin{equation}
p_r(\phi_0)=\mathbb{P}\left(\exists\, x\in \mathcal{B}_r^1:\phi(x)=0\,|\,\phi(x_0)=\phi_0\right)\,.
\end{equation}
Conditioning on $\phi(x_0) = \phi_0$, $\phi$ becomes the Gaussian process with average
\begin{equation}
\mu(x) = K(x,x_0)\,\phi_0\left/K(x_0,x_0)\right.
\end{equation}
and covariance
\begin{equation}
\hat{K}(x,y) = K(x,y) - \frac{K(x,x_0)\,K(x_0,y)}{K(x_0,x_0)}\,.
\end{equation}
We put $\phi(x) = \mu(x) - \varphi(x)$ for any $x\in\mathbb{R}^n$, such that $\varphi$ is a centered Gaussian process with covariance $\hat{K}$.
Let
\begin{align}
K_r &= \inf_{x\in \mathcal{B}_r^1}\frac{K(x,x_0)}{K(x_0,x_0)}\,,\nonumber\\
\varphi_r &= \mathbb{E}\sup_{x\in \mathcal{B}_r^1}\varphi(x)\,,\qquad\sigma_r^2=\sup_{x\in \mathcal{B}_r^1}\hat{K}(x,x)\,,
\end{align}
and let us assume that $K_r\,\phi_0>\varphi_r$.
The Borell--TIS inequality \cite{adler2009random} provides an upper bound to $p_r(\phi_0)$:
\begin{thm*}[Borell--TIS inequality]\label{thm:TIS}
Let $\varphi$ be a centered Gaussian process on $\Omega\subset\mathbb{R}^n$, and let $\hat{K}$ be the associated kernel.
Then, for any $t>0$
\begin{equation}
\mathbb{P}\left(\sup_{x\in \Omega}\varphi(x)\ge \mathbb{E}\sup_{x\in \Omega}\varphi(x) + t\right) \le \mathrm{e}^{-\frac{t^2}{2\,\sigma^2}}\,,
\end{equation}
where $\sigma^2=\sup_{x\in \Omega}\hat{K}(x,x)$.
\end{thm*}
We have from the Borell--TIS inequality
\begin{align}\label{eq:pr0}
&p_r(\phi_0) \le \mathbb{P}\left(\exists\, x\in \mathcal{B}_r^1:\varphi(x)\ge\mu(x)\right)\nonumber\\
&\le \mathbb{P}\left(\exists\, x\in \mathcal{B}_r^1:\varphi(x)\ge K_r\,\phi_0\right) \le \mathrm{e}^{-\frac{\left(K_r\,\phi_0 - \varphi_r\right)^2}{2\,\sigma_r^2}}\,.
\end{align}
Recalling that $\phi(x_0)$ is a centered Gaussian random variable with variance $K(x_0,x_0)$, we have
\begin{align}\label{eq:pr}
p_r &= 2\int_0^\infty p_r(\phi_0)\,\mathrm{e}^{-\frac{\phi_0^2}{2\,K(x_0,x_0)}}\,\frac{\mathrm{d}\phi_0}{\sqrt{2\pi\,K(x_0,x_0)}}\nonumber\\
&\le 2\int_0^\frac{\varphi_r}{K_r}\mathrm{e}^{-\frac{\phi_0^2}{2\,K(x_0,x_0)}}\,\frac{\mathrm{d}\phi_0}{\sqrt{2\pi\,K(x_0,x_0)}}\nonumber\\
&\phantom{\le} + 2\int_\frac{\varphi_r}{K_r}^\infty\mathrm{e}^{-\frac{\left(K_r\,\phi_0 - \varphi_r\right)^2}{2\,\sigma_r^2}-\frac{\phi_0^2}{2\,K(x_0,x_0)}}\,\frac{\mathrm{d}\phi_0}{\sqrt{2\pi\,K(x_0,x_0)}}\nonumber\\
&\le \frac{\sqrt{\frac{2}{\pi}}\,\varphi_r + \sigma_r}{K_r\sqrt{K(x_0,x_0)}}\,.
\end{align}
We get an upper bound on $\varphi_r$ from Dudley's theorem \cite{bartlett2013theoretical}.
\begin{thm*}[Dudley's theorem]
Let $\varphi$ be a centered Gaussian process on $\Omega\subset\mathbb{R}^n$, and let $\hat{d}$ be the RKHS distance of the associated kernel.
For any $\epsilon>0$, let $N(\epsilon)$ be the minimum number of balls of $\hat{d}$ with radius $\epsilon$ that can cover $\Omega$.
Then,
\begin{equation}
\mathbb{E}\sup_{x\in\Omega}\varphi(x) \le 8\sqrt{2}\int_0^\infty\sqrt{\ln N(\epsilon)}\,\mathrm{d}\epsilon\,.
\end{equation}
\end{thm*}
We directly get from Dudley's theorem
\begin{equation}\label{eq:Dudley}
\varphi_r \le 8\sqrt{2}\int_0^\infty\sqrt{\ln N_r(\epsilon)}\,\mathrm{d}\epsilon\,,
\end{equation}
where $N_r(\epsilon)$ is the minimum number of balls of $\hat{d}$ with radius $\epsilon$ that can cover $\mathcal{B}_r^1$.
Let $N(\epsilon)$ be the minimum number of balls of the Euclidean distance with radius $\epsilon$ that can cover the unit $\ell^1$ ball.
In \autoref{sec:lemdhproof} we prove the following \autoref{lem:dh}:
\begin{lem}\label{lem:dh}
$\hat{d}(x,y)\le d(x,y)$ for any $x,\,y\in\mathbb{R}^n$.
\end{lem}
From \autoref{lem:dh} and \eqref{eq:C12} we get $\hat{d}(x,y) \le M\,C\left\|x-y\right\|_2$ for any $x,\,y\in\mathbb{R}^n$, therefore $N_r(\epsilon) \le N\left(\epsilon/(M\,C\,r)\right)$ and \eqref{eq:Dudley} implies
\begin{equation}\label{eq:phir}
\varphi_r \le 8\sqrt{2}\,M\,C\,r\int_0^\infty\sqrt{\ln N(\epsilon)}\,\mathrm{d}\epsilon\,.
\end{equation}
In \autoref{sec:thmNepsilonproof} we prove the following \autoref{thm:Nepsilon}:
\begin{thm}\label{thm:Nepsilon}
For any $\epsilon>0$, the open unit ball $\mathcal{B}_1$ of the $\ell^1$ norm in $\mathbb{R}^n$ can be covered with
\begin{equation}
N(\epsilon) \le \left\{
                  \begin{array}{ll}
                    1\quad& \epsilon\ge1\\
                    \left(2n\right)^\frac{1}{\epsilon^2} \quad& \frac{1}{\sqrt{n}}<\epsilon<1 \\
                    \left(1+\frac{2}{\epsilon}\right)^n\quad& 0<\epsilon\le\frac{1}{\sqrt{n}}
                  \end{array}
                \right.
\end{equation}
balls of the Euclidean distance with radius $\epsilon$ and centers in $\mathcal{B}_1$.
\end{thm}
We get from \autoref{thm:Nepsilon}
\begin{align}\label{eq:cn}
&\int_0^\infty\sqrt{\ln N(\epsilon)}\,\mathrm{d}\epsilon \le \nonumber\\
&\int_0^\frac{1}{\sqrt{n}}\sqrt{n\ln\left(1+\frac{2}{\epsilon}\right)}\,\mathrm{d}\epsilon + \sqrt{\ln2n}\int_\frac{1}{\sqrt{n}}^1\frac{\mathrm{d}\epsilon}{\epsilon}\nonumber\\
&= \sqrt{\frac{\ln 4n}{2}}\int_0^1\sqrt{\ln\left(\frac{1}{2\sqrt{n}} + \frac{1}{x}\right)}\,\mathrm{d}x + \frac{\ln n}{2}\sqrt{\ln 2n}\nonumber\\
&\le \sqrt{\frac{\ln 4n}{2}}\int_0^1\sqrt{\ln\left(\frac{1}{2\sqrt{2}} + \frac{1}{x}\right)}\,\mathrm{d}x + \frac{\ln n}{2}\sqrt{\ln 2n}\nonumber\\
&\le \frac{3}{4}\sqrt{\ln 4n} + \frac{\ln n}{2}\sqrt{\ln 2n} = a_n\,,
\end{align}
where in the second line we made the change of variable $x = \epsilon\sqrt{n}$.
In \autoref{sec:lemKrproof} and \autoref{sec:lemsigmarproof}, we prove the following \autoref{lem:Kr} and \autoref{lem:sigmar}, respectively:
\begin{lem}\label{lem:Kr}
We have
\begin{equation}
K_r\ge 1 - M\,C\,r\left/\sqrt{K(x_0,x_0)}\right.\,.
\end{equation}
\end{lem}
\begin{lem}\label{lem:sigmar}
We have $\sigma_r \le M\,C\,r$.
\end{lem}
Putting together \eqref{eq:cn}, \eqref{eq:phir}, \eqref{eq:pr}, \autoref{lem:Kr} and \autoref{lem:sigmar} we get
\begin{equation}\label{eq:prineq}
p_r \le \frac{\frac{16}{\sqrt{\pi}}\,a_n + 1}{\frac{\sqrt{K(x_0,x_0)}}{M\,C\,r} - 1} \le \frac{\frac{16}{\sqrt{\pi}}\,a_n + 1}{\frac{\left\|x_0\right\|_2}{M\,r} - 1}\,,
\end{equation}
where the last inequality follows from \eqref{eq:C12}.
Therefore, we have $p_r\le\delta$ for
\begin{equation}
    \frac{M\,r}{\left\|x_0\right\|_2} \le \frac{\delta}{\frac{16}{\sqrt{\pi}}\,a_n + 1 + \delta}\,,
\end{equation}
and the claim follows.

\section{Experiments}
\label{sec:adv_attack_random}

\begin{figure*}[t!]
\centering
  \includegraphics[width=0.8\textwidth]{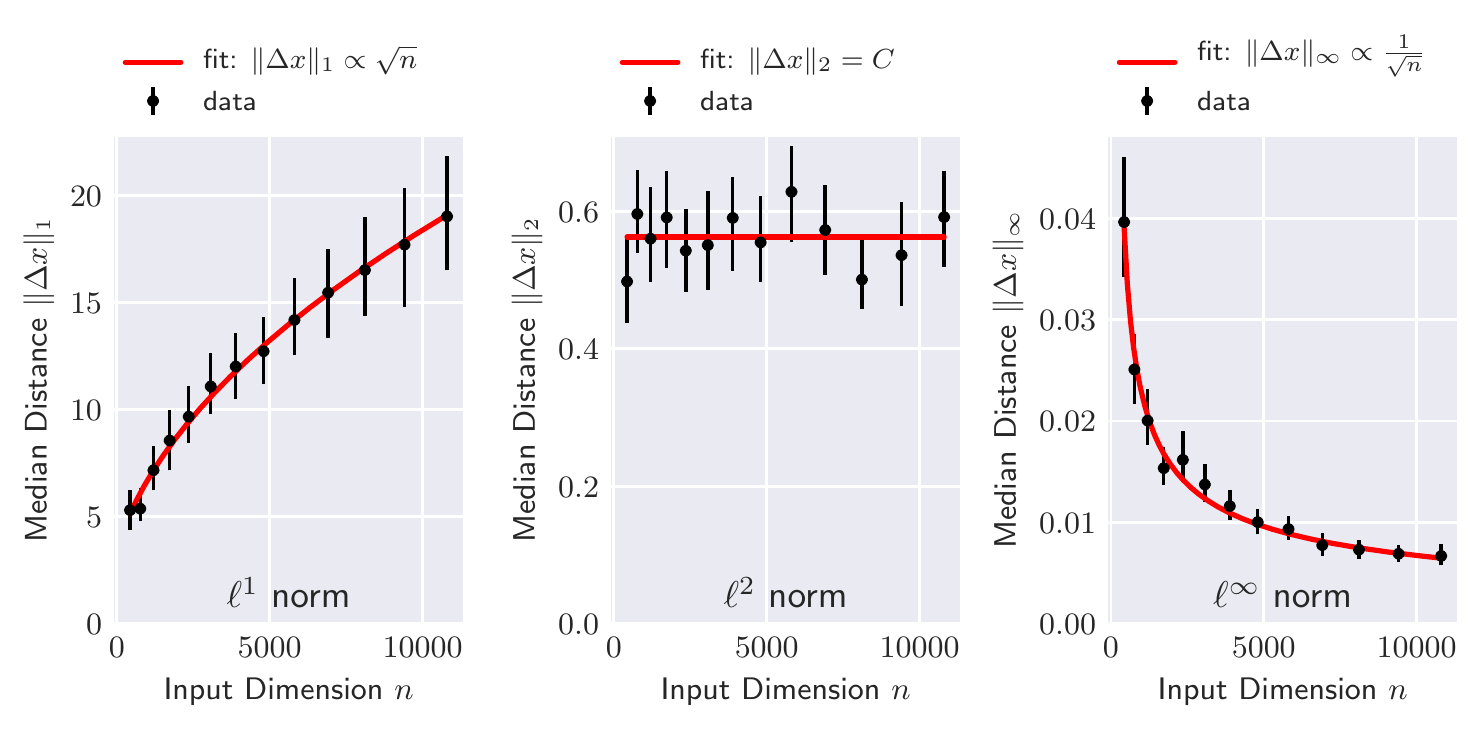}
  \caption{{\bf Random untrained residual networks:} The median $\ell^p$ distances of closest adversarial examples from their respective inputs for $p =  1,\,2,\,\infty$ scale as predicted in \autoref{rem:norm_p_scaling} for a residual network (see \autoref{app:net_architectures} for full description of network). Error bars span $\pm 5$ percentiles from the median. For each input dimension, results are calculated from 2000 samples (200 random networks each attacked at 10 random points). See \autoref{app:attack_methods} for further details on how experiments were performed. }\label{fig:random_resnet}
\end{figure*}

To experimentally validate \autoref{cor:main} and \autoref{rem:norm_p_scaling}, we performed adversarial attacks on random inputs for various network architectures with randomly chosen weights \footnote{code to replicate experiments published at \href{https://github.com/bkiani/Adversarial-robustness-guarantees-for-random-deep-neural-networks}{https://github.com/bkiani/Adversarial-robustness-guarantees-for-random-deep-neural-networks}}.
As shown in \autoref{app:supplementary_figures}, experimental findings were consistent across a variety of networks.
For sake of brevity, in this section we only provide figures and results for a simplified residual network. \autoref{fig:random_resnet} plots the median distance of adversarial examples for a residual network similar to the first proposed residual network \cite{he2016deep}. This network contains three residual blocks and does not contain a global average pooling layer before the final output (its complete architecture is given in \autoref{app:net_architectures}). Attacks were performed on 2-dimensional images with three channels and pixel values chosen randomly from the standard uniform distribution.

\begin{figure}[!ht]
  \centering
  \includegraphics[width=\columnwidth]{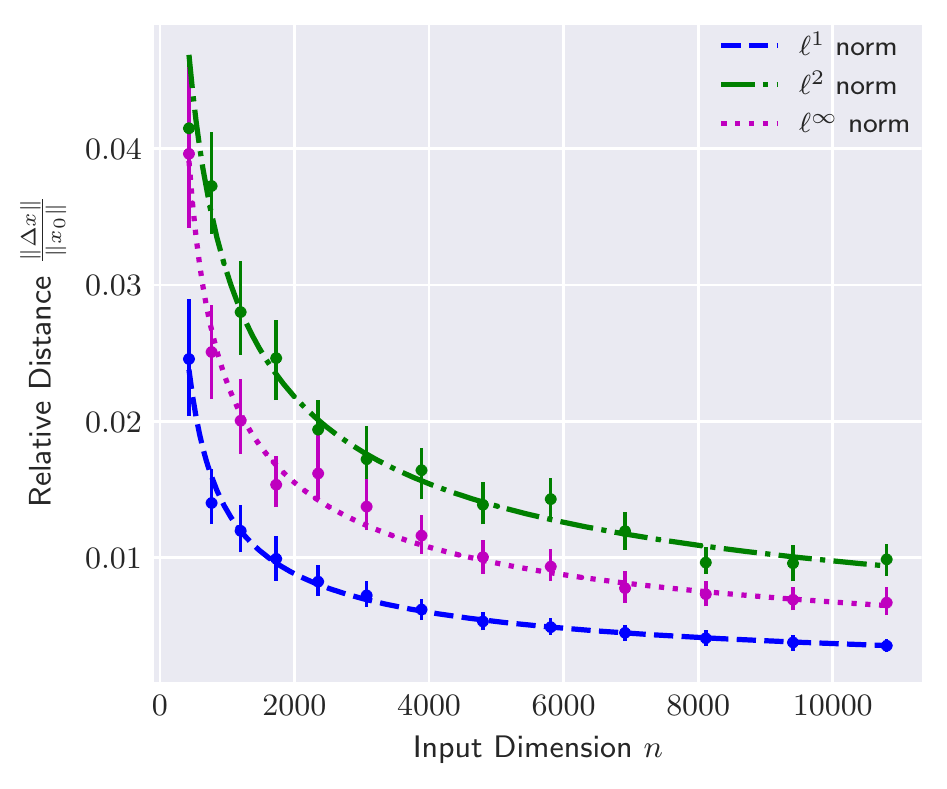}
  \caption{{\bf Random untrained networks:} Median relative distance of closest adversarial examples $\| \Delta x \|_p / \| x_0 \|_p$ from their respective inputs ($p \in \{1,\,2,\,\infty\}$) scale with the input dimension $n$ as $O(1/\sqrt{n})$ in all norms for a residual network with random weights (see \autoref{app:net_architectures} for full description of network), confirming the theoretical predictions of \autoref{rem:norm_p_scaling}. Results plotted here are for residual networks with random weights. Error bars span $\pm 5$ percentiles from the median. For each input dimension, results are calculated from 2000 samples (200 random networks each attacked at 10 random points).
  }
  \label{fig:random_resnet_normalized}
\end{figure}

Results from \autoref{fig:random_resnet} plotting median adversarial distances as a function of the input dimension are consistent with the expected theoretical scaling in \autoref{rem:norm_p_scaling}. Namely, adversarial distances in the $\ell^1$, $\ell^2$, and $\ell^\infty$ norms scale with the dimension of the input $n$ proportionally to $\sqrt{n}$, a constant $C$ (not dependent on $n$), and $1/\sqrt{n}$, respectively (up to logarithmic factors). Adversarial distances relative to the average starting norm of an input are plotted in \autoref{fig:random_resnet_normalized}. This adjusted metric named relative distance provides a convenient means of understanding the scaling of adversarial distances, since relative adversarial distances scale proportionally to $1/\sqrt{n}$ in all norms.

\subsection{Adversarial Attacks on Trained Neural Networks}
\label{sec:adv_attack_trained}

\begin{figure*}[t!]
\centering
  \includegraphics[]{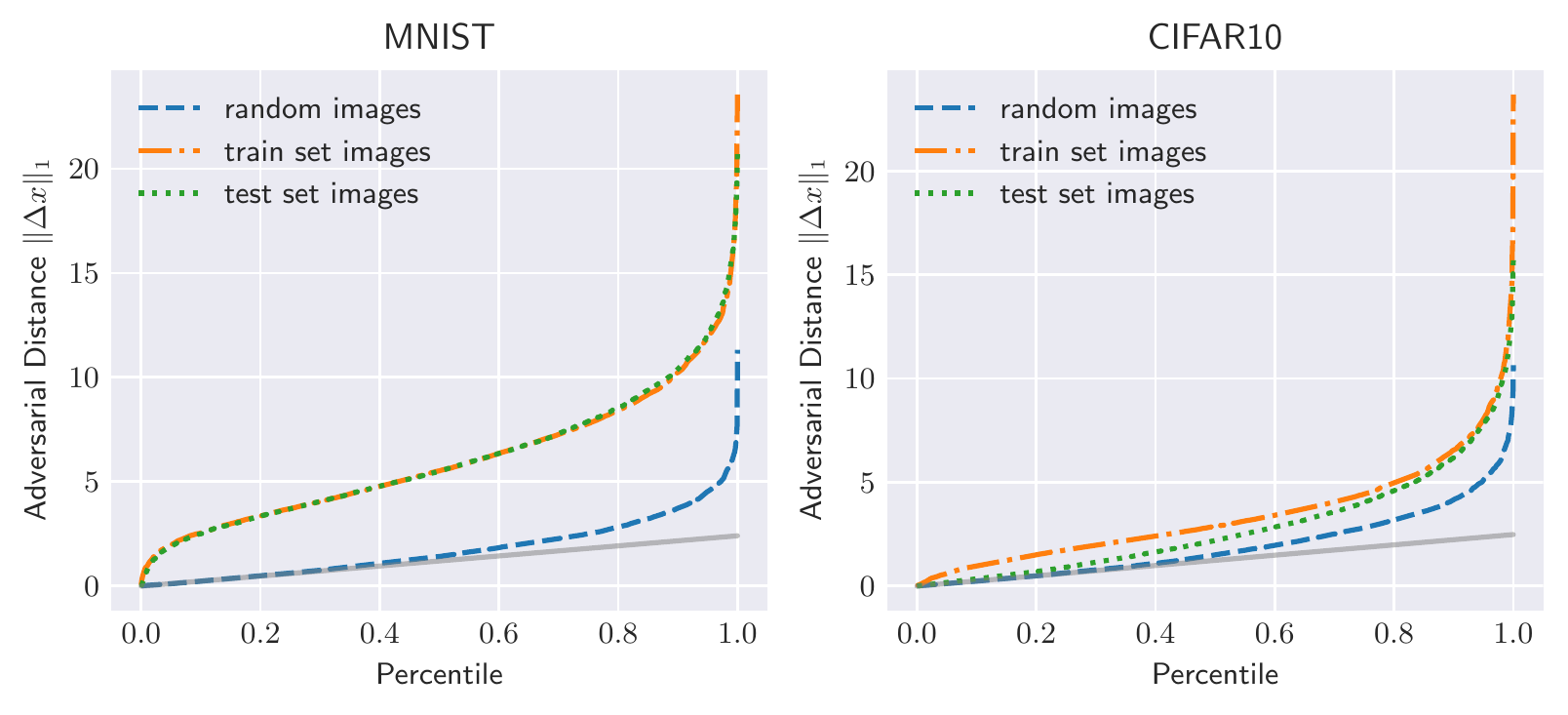}
  \caption{{\bf Trained networks:} Adversarial distance by percentile for random images (images with randomly chosen pixel values) and images in the training and test sets. The expected linear relationship between distance and percentile is observed for random images apart from the highest percentiles as is evident from the linear fit over percentiles ranging from 0 to 0.25 shown as solid line. Adversarial attacks are performed on the $\ell^1$ norm. Network architecture is a simplified residual network (see \autoref{app:trained_networks}).}
  \label{fig:trained_resnet_profile}
\end{figure*}

Results from \autoref{sec:adv_attack_random} indicate that adversarial attacks on networks with randomly chosen weights empirically conform with our main findings presented in \autoref{sec:main_results}. In this section, we extend our experimental analysis to networks trained on MNIST and CIFAR10 data. We trained networks with the same residual network architecture given in the prior section on MNIST and CIFAR10 data under the task of binary classification.
For the case of MNIST, the binary classification task was determining if a digit is odd or even. For CIFAR10, image classes were assigned to binary categories of either \{airplane, bird, deer, frog, ship\} or \{automobile, cat, dog, horse, truck\}.
Networks were trained for 15 and 25 epochs for the MNIST and CIFAR10 datasets respectively achieving greater than 98\% training set accuracy in all cases. We refer to \autoref{app:trained_networks} for full details on the training of the networks.

\begin{figure*}[ht]
\centering
  \includegraphics[]{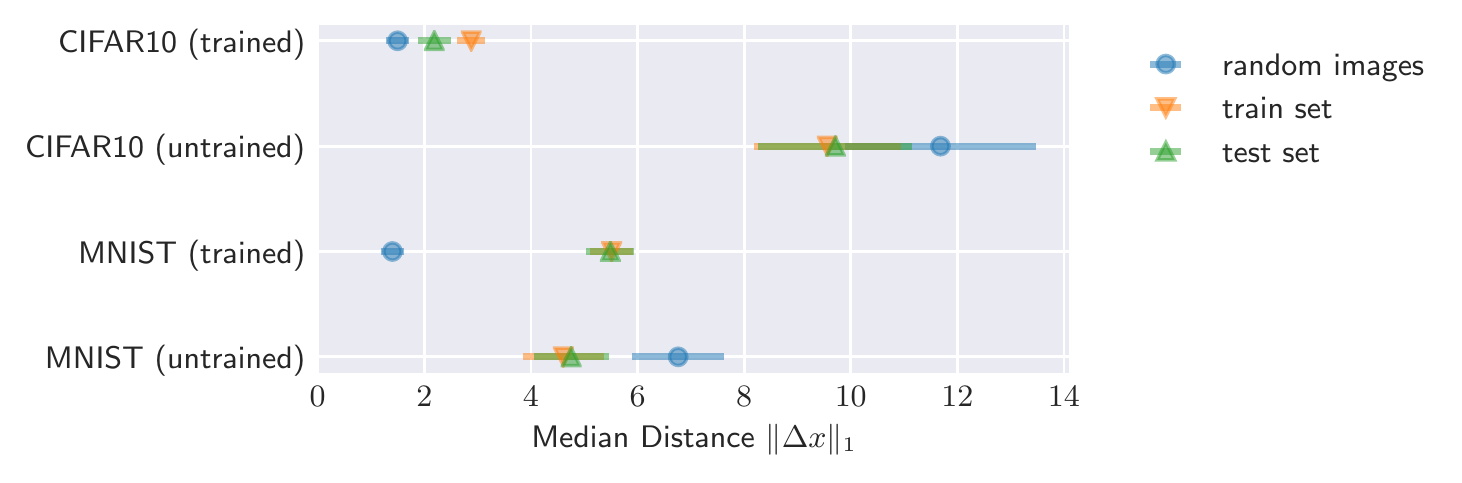}
  \caption{{\bf Random vs trained networks:} Median distance of adversarial examples (in $\ell^1$ norm) for random neural networks and neural networks of same architecture trained on MNIST and CIFAR10 data \autoref{fig:random_resnet}. Analysis is performed for random images (images with randomly chosen pixel values) and images in the training and test sets. Network architecture is a simplified residual network (see \autoref{app:net_architectures}).}
  \label{fig:trained_resnet_comparison}
\end{figure*}

Properties of trained neural networks, especially as they relate to adversarial robustness and generalization, are dependent on the properties of the data used to train them.
For example, since neural networks can be trained to ``memorize'' data \cite{choromanska2015loss}, \autoref{cor:main} can be forced to fail if the network is trained on a dataset which contains very close inputs with different labels.
From \autoref{fig:trained_resnet_comparison}, the networks trained on CIFAR10 data show a smaller adversarial distance with respect to random networks on both random images and images taken from the training or test set.
In the case of MNIST, training decreases the adversarial distance for random images, but does not significantly change it for training or test images.
A possible explanation for this discrepancy is the conspicuous geometric and visual structure inherent in the MNIST dataset relative to CIFAR10. Digits in MNIST all have the same uniform black background and geometry and roughly fill the whole image, while in CIFAR10 the background and the relative size of the relevant part of the image can vary significantly, and pictures are taken from various different angles (\emph{e.g.,} different orientations of a dog or car). Thus, when trained on MNIST, networks can more easily embed training and test points within areas far from classification boundaries. More generally, networks trained on MNIST data achieve low generalization error and increased adversarial distances are correlated with those lower errors (though adversarial robustness can sometimes be at odds with generalization \cite{raghunathan2019adversarial,tsipras2018robustness}).
On the other hand, the networks trained on CIFAR10 slightly suffer from overfitting, since they have a $16.3\%$ discrepancy between the performances on the training and the test data.
Therefore, the lower adversarial distance might be due to noise-like signals employed for prediction.

Another possible explanation of the discrepancy in the size of the adversarial perturbations between random and trained deep neural networks is that networks trained with stochastic gradient descent are known to be less robust to adversarial perturbations than networks trained with Bayesian inference \cite{duvenaud2016early,bekasov2018bayesian,carbone2020robustness}.
As shown in \autoref{sec:intro}, under the hypothesis that the target function is generated by a given random deep neural network, the classifier obtained from Bayesian training of the same network has the same properties as a function generated by the random network.
Therefore we expect that, for the size of the adversarial perturbations, random deep neural networks are closer to deep neural networks trained with Bayesian inference than to  deep neural networks trained with gradient descent.

From \autoref{cor:main}, we expect the portion of images that have at least one adversarial example within a given $\ell^1$ distance to increase linearly with the distance.
This finding is validated by results shown in \autoref{fig:trained_resnet_profile} which plots the adversarial distance by percentile (sorted smallest to largest distance). In the case of random images, the linear increase in adversarial distance by percentile is evident throughout most of the percentiles in the chart conforming closely to the linear fit (dotted line). Interestingly, this linear correlation is even observed in images in the training and test sets outside of the smallest and highest percentiles. For training and test set images, networks usually predicted labels with high confidence thus limiting the percentage of images falling at small distances from a classification boundary.

\section{Discussion}
We have studied the properties of adversarial examples for deep neural networks with random weights and biases and have proved that for any $p\ge1$, the $\ell^p$ distance from the closest classification boundary of any given input is with high probability at least $\tilde{\Omega}\left(1/\sqrt{n}\right)$ times the $\ell^p$ norm of the input, where $n$ is the dimension of the input space (\autoref{cor:main} and \autoref{rem:norm_p_scaling}).
This lower bound matches the upper bound of \cite{gilmer2018adversarial,fawzi2018adversarial,shafahi2019adversarial,mahloujifar2019curse}, and our result determines the universal scaling of the minimum size of adversarial perturbations.
Under the hypothesis that the target function is generated by a given random deep neural network, our probabilistic robustness guarantee also applies to the same network trained with Bayesian inference.

We have validated our theoretical results with experiments on both random deep neural networks and deep neural networks trained with stochastic gradient descent on the MNIST and CIFAR10 datasets.
The experiments on random networks are in complete agreement with our theoretical predictions.
Networks trained on MNIST and CIFAR10 data are mostly consistent with our main findings, and we conjecture that the proof of our adversarial robustness guarantee can be extended to trained deep neural networks.
Indeed, \cite{jacot2018neural,lee2019wide,yang2019scaling,arora2019exact,huang2019dynamics,li2019enhanced,wei2019regularization,cao2019generalization} have proved that the training of deep neural networks via stochastic gradient descent is similar to the training of Gaussian processes with Bayesian inference, and the classifier generated by a trained deep neural network still behaves as a Gaussian process.
Therefore, we expect that the robustness of deep neural networks trained with stochastic gradient descent can be studied with similar techniques, and we believe that a probabilistic robustness guarantee similar to the guarantee proved in this paper can be proved.
While the robustness guarantee for random deep neural networks does not require any assumption on the inputs, extensions to trained deep neural networks will definitely require assumptions on the training data.
Given our results on random untrained networks extend directly to random networks trained via Bayesian inference under the hypothesis that the target classifier is drawn from the same random distribution, we conjecture that extending our results more broadly to trained networks will require that training labels look ``typical" with respect to the probability distribution generated by the random initialization of the deep neural network at the beginning of the training.

The robustness guarantee of \autoref{cor:main} depends on the architecture of the deep neural network through the ratio between the number of pixels of the output and of the input layer, favoring the case where such ratio is not small.
We expect that with similar techniques the dependence of the bound on the architecture can be refined, thus allowing for a systematic study of how the choice of the model affects the adversarial robustness.
Moreover, the extension of our results to deep neural networks trained with stochastic gradient descent would provide an analytic lower bound to the size of adversarial perturbations in terms of their architecture, and would therefore open the way to the first thorough theoretical understanding of the relationship between the network architecture and its robustness to adversarial attacks.

Finally, our methods can be employed to study the robustness of deep neural networks with respect to adversarial perturbations that keep the data manifold invariant, such as smooth deformations of the input image \cite{mallat2012group,oyallon2015deep,bruna2013invariant,bietti2019group,bietti2019inductive}.

\section*{Acknowledgements}

We thank Milad Marvian, Dario Trevisan and Laurent B\'etermin for useful discussions.

This work was supported by the USA Air Force Office of Scientific Research, the USA Army Research Office under the Blue Sky program, DOE and IARPA.

\appendix
\onecolumn

\section{Proof of \autoref{thm:main}, Part II}\label{sec:mainproofII}
Let
\begin{equation}
p_r = \mathbb{P}\left\{\exists\, x\in \mathcal{L}_r : \phi(x)=0\right\}\,.
\end{equation}
We define for any $0\le t\le r$
\begin{equation}
f(t) = \frac{\phi(x_0+t\,v)}{\sqrt{K(x_0+t\,v,\,x_0+t\,v)}}\,,
\end{equation}
such that $f$ is a centered Gaussian process on $[0,r]$ with covariance and feature map
\begin{equation}
\tilde{K}(s,t) = \frac{K(x_0+s\,v,\,x_0+t\,v)}{\sqrt{K(x_0+s\,v,\,x_0+s\,v)\,K(x_0+t\,v,\,x_0+t\,v)}}\,,\qquad \tilde{\Phi}(t) = \frac{\Phi(x_0 + t\,v)}{\left\|\Phi(x_0 + t\,v)\right\|}\,.
\end{equation}
We have
\begin{align}
\left\|\tilde{\Phi}(s) - \tilde{\Phi}(t)\right\| &= \frac{\left\|\left\|\Phi(x_0+t\,v)\right\|\Phi(x_0+s\,v) - \left\|\Phi(x_0+s\,v)\right\|\Phi(x_0+t\,v)\right\|}{\left\|\Phi(x_0+s\,v)\right\|\left\|\Phi(x_0+t\,v)\right\|}\nonumber\\
&\le \frac{\left\|\Phi(x_0+s\,v)-\Phi(x_0+t\,v)\right\| + \left|\left\|\Phi(x_0+s\,v)\right\| - \left\|\Phi(x_0+t\,v)\right\|\right|}{\left\|\Phi(x_0+s\,v)\right\|}\nonumber\\
&\le \frac{2\,M\left|s-t\right|}{\left\|x_0+s\,v\right\|_2} \le \frac{2\,M\left|s-t\right|}{\left\|x_0\right\|_2-r}\,.
\end{align}
To conclude, we will need the following two results.
The first is Rice's formula, which provides an upper bound to the number of zeroes of a one-dimensional Gaussian process on a given interval:
\begin{thm*}[Rice's formula \cite{adler2009random}]
Let $f:[0,r]\to\mathbb{R}$ be a centered Gaussian process with covariance $\tilde{K}$ such that $\tilde{K}(t,t)=1$ for any $0\le t\le r$.
Then,
\begin{equation}
\mathbb{E}\left|\left\{0\le t\le r : f(t) = 0\right\}\right| = \frac{1}{\pi}\int_0^r\sqrt{\left.\frac{\partial^2}{\partial s\partial t}\tilde{K}(s,t)\right|_{s=t}}\,\mathrm{d}t\,.
\end{equation}
\end{thm*}
The second is the following \autoref{lem:f'}:
\begin{lem}\label{lem:f'}
Let $\tilde{d}$ be the distance associated with $\tilde{K}$, and let us assume that for any $s,\,t\in[0,r]$
\begin{equation}
\tilde{d}(s,t) \le \tilde{C}\left|s-t\right|\,.
\end{equation}
Then,
\begin{equation}
\left.\frac{\partial^2}{\partial s\partial t}\tilde{K}(s,t)\right|_{s=t} \le {\tilde{C}}^2\,.
\end{equation}
\begin{proof}
Let $\tilde{\Phi}$ be the feature map associated with $\tilde{K}$.
We have
\begin{align}
\left.\frac{\partial^2}{\partial s\partial t}\tilde{K}(s,t)\right|_{s=t} &= \lim_{\epsilon\to 0}\frac{\left(\tilde{\Phi}(s+\epsilon)-\tilde{\Phi}(s)\right)\cdot\left(\tilde{\Phi}(t+\epsilon)-\tilde{\Phi}(t)\right)}{\epsilon^2}\nonumber\\
&\le \lim_{\epsilon\to 0}\frac{\left\|\tilde{\Phi}(s+\epsilon)-\tilde{\Phi}(s)\right\|\left\|\tilde{\Phi}(t+\epsilon)-\tilde{\Phi}(t)\right\|}{\epsilon^2} \le {\tilde{C}}^2\,.
\end{align}
\end{proof}
\end{lem}
Rice's formula and \autoref{lem:f'} imply
\begin{equation}
p_r \le \mathbb{E}\left|\left\{0\le t\le r : \phi(x_0 + t\,v) = 0\right\}\right| \le \frac{2\,M\,r}{\pi\left(\left\|x_0\right\|_2-r\right)}\,,
\end{equation}
and the claim follows.

\section{Proof of \autoref{thm:DNN}}\label{sec:DNNproof}
The proof of \autoref{thm:DNN} is based on the following theorem, which formalizes the equivalence between deep neural networks with random weights and biases and Gaussian processes.
\begin{thm}[Master Theorem \cite{yang2019wide}]\label{thm:master}
Let $\phi^{(1)},\,\ldots,\,\phi^{(L+1)}$ be the outputs of the layers of the random deep neural network defined in \autoref{sec:setup}.
Let $K^{(1)},\,\ldots,\,K^{(L+1)}$ be the kernels on $\mathbb{R}^{n_C^{(0)}\times\mathcal{I}^{(0)}}$ where $K^{(l)}$ is recursively defined as
\begin{equation}\label{eq:defKl}
\mathbb{E}\left(\phi^{(l)}_{i,\alpha}(x)\,\phi^{(l)}_{j,\beta}(y)\right) = \delta_{ij}\,K^{(l)}_{\alpha,\beta}(x,y)\,,\qquad i,\,j=1,\,\ldots,\,n_C^{(l)}\,,\quad\alpha,\,\beta\in\mathcal{I}^{(l)}\,,
\end{equation}
i.e., as the covariance of $\phi^{(l)}$, where the expectation is computed assuming that $\phi^{(1)},\,\ldots,\,\phi^{(l-1)}$ are independent centered Gaussian processes with covariances $K^{(1)},\,\ldots,\,K^{(l-1)}$.

Given $M\in\mathbb{N}$, let $\psi:\mathbb{R}^M\to\mathbb{R}$ be such that there exist $A,\,a,\,\epsilon>0$ such that for any $z\in\mathbb{R}^M$,
\begin{equation}\label{eq:psi}
|\psi(z)|\le \exp\left(A\left\|z\right\|_2^{2-\epsilon}+a\right)\,.
\end{equation}
Then, in the limit $n_C^{(1)},\,\ldots,\,n_C^{(L+1)}\to\infty$, we have for any $x^1,\,\ldots,\,x^M\in\mathbb{R}^{n_C^{(0)}\times\mathcal{I}^{(0)}}$
\begin{equation}
\frac{1}{n_C^{(L+1)}}\sum_{i=1}^{n_C^{(L+1)}}\psi\left(\phi^{(L+1)}_i\left(x^1\right),\,\ldots,\,\phi^{(L+1)}_i\left(x^M\right)\right) \overset{\mathrm{a.s.}}{\longrightarrow}\mathbb{E}_{Z\sim\mathcal{N}(0,\Sigma)}\,\psi(Z)\,,
\end{equation}
with the covariance matrix $\Sigma$ given by
\begin{equation}\label{eq:Sigma}
\Sigma_{mm'} = K^{(L+1)}\left(x^m,x^{m'}\right)\,,\qquad m,\,m'=1,\,\ldots,\,M\,.
\end{equation}
\end{thm}
\begin{rem}
For finite width, the outputs of the intermediate layers of the random deep neural networks have a sub-Weibull distribution \cite{vladimirova2019understanding}.
\end{rem}
The main consequence of \autoref{thm:master} is that the final output $\phi$ is a centered Gaussian process:
\begin{cor}
The final output of the deep neural network $\phi$ is a centered Gaussian process with covariance $K=K^{(L+1)}$.
\begin{proof}
Given $M\in\mathbb{N}$, let $\psi:\mathbb{R}^M\to\mathbb{R}$ be continuous and bounded.
For any $x^1,\,\ldots,\,x^M\in\mathbb{R}^{n_C^{(0)}\times\mathcal{I}^{(0)}}$ we have from \autoref{thm:master}
in the limit $n_C^{(1)},\,\ldots,\,n_C^{(L+1)}\to\infty$
\begin{equation}\label{eq:as}
\frac{1}{n_C^{(L+1)}}\sum_{i=1}^{n_C^{(L+1)}}\psi\left(\phi^{(L+1)}_i\left(x^1\right),\,\ldots,\,\phi^{(L+1)}_i\left(x^M\right)\right) \overset{\mathrm{a.s.}}{\longrightarrow}\mathbb{E}_{Z\sim\mathcal{N}(0,\Sigma)}\,\psi(Z)\,,
\end{equation}
with $\Sigma$ as in \eqref{eq:Sigma}.
Taking the expectation value on both sides of \eqref{eq:as} we get, recalling that each $\phi^{(L+1)}_i$ has the same probability distribution as $\phi$,
\begin{equation}
\lim_{n_C^{(1)},\,\ldots,\,n_C^{(L)}\to\infty}\mathbb{E}\,\psi\left(\phi\left(x^1\right),\,\ldots,\,\phi\left(x^M\right)\right) = \mathbb{E}_{Z\sim\mathcal{N}(0,\Sigma)}\,\psi(Z)\,,
\end{equation}
and the claim follows.
\end{proof}
\end{cor}

It is convenient to define for any $l=1,\,\ldots,\,L$
\begin{equation}
K^{(l)}(x,y) = \sum_{\alpha\in\mathcal{I}^{(l)}}K^{(l)}_{\alpha,\alpha}(x,y)\,.
\end{equation}
Let also
\begin{equation}
{d^{(l)}(x,y)}^2 = K^{(l)}(x,x) -2\,K^{(l)}(x,y) + K^{(l)}(y,y)
\end{equation}
be the RKHS distance associated with $K^{(l)}$.

We will prove by induction that for any $l=1,\,\ldots,\,L+1$ there exist $C^{(l)},\,M^{(l)}>0$ such that $K^{(l)}$ satisfies \eqref{eq:C12} with $C=C^{(l)}$ and $M=M^{(l)}$.
The following subsections will prove the inductive step for each of the types of layer defined in \autoref{sec:setup}.

\subsection{Input Layer}
$\phi^{(1)}(x)$ is a centered Gaussian process with covariance as in \eqref{eq:defKl} with
\begin{equation}
K^{(1)}_{\alpha,\beta}(x,y) = {\sigma^{(1)}_b}^2 + {\sigma^{(1)}_W}^2\sum_{i=1}^{n_C^{(0)}}\sum_{\gamma\in\mathcal{P}^{(1)}}\frac{x_{i,\alpha+\gamma}\,y_{i,\beta+\gamma}}{n_C^{(0)}}\,,\qquad x,\,y\in\mathbb{R}^{n_C^{(0)}\times\mathcal{I}^{(0)}},\quad\alpha,\,\beta\in\mathcal{I}^{(1)}\,,
\end{equation}
and
\begin{equation}
K^{(1)}(x,y) = \left|\mathcal{I}^{(1)}\right|{\sigma^{(1)}_b}^2 + \frac{\left|\mathcal{P}^{(1)}\right|{\sigma^{(1)}_W}^2}{n_C^{(0)}}\,x\cdot y\,,
\end{equation}
therefore
\begin{align}
{d^{(1)}(x,y)}^2 &= \frac{\left|\mathcal{P}^{(1)}\right|{\sigma^{(1)}_W}^2}{n_C^{(0)}}\left\|x-y\right\|_2^2\,,\nonumber\\
K^{(1)}(x,x) &= \left|\mathcal{I}^{(1)}\right|{\sigma^{(1)}_b}^2 + \frac{\left|\mathcal{P}^{(1)}\right|{\sigma^{(1)}_W}^2}{n_C^{(0)}}\left\|x\right\|_2^2\ge \frac{\left|\mathcal{P}^{(1)}\right|{\sigma^{(1)}_W}^2}{n_C^{(0)}}\left\|x\right\|_2^2\,,
\end{align}
and $K^{(1)}$ satisfies \eqref{eq:C12} with
\begin{equation}
C^{(1)}={\sigma^{(1)}_W}\sqrt{\frac{\left|\mathcal{P}^{(1)}\right|}{n_C^{(0)}}}\,,\qquad M^{(1)}=\sqrt{\frac{\left|\mathcal{I}^{(0)}\right|}{\left|\mathcal{I}^{(1)}\right|}}=1\,.
\end{equation}

\subsection{Nonlinear Layer}\label{sec:conv}
Let the ($l+1$)-th layer be a nonlinear layer.
From \autoref{thm:master}, we can assume that $\phi^{(l)}$ is the centered Gaussian process with covariance $K^{(l)}$.
We then have
\begin{equation}\label{eq:Kl+1tau}
K^{(l+1)}_{\alpha,\beta}(x,y) = {\sigma^{(l+1)}_b}^2 + {\sigma^{(l+1)}_W}^2\sum_{\gamma\in\mathcal{P}^{(l+1)}}\mathbb{E}\left(\tau(u)\,\tau(v):(u,v)\sim\mathcal{N}\left(0,\Sigma^{(l)}_{\alpha,\beta}(x,y)\right)\right)\,,
\end{equation}
with
\begin{equation}
\Sigma^{(l)}_{\alpha,\beta}(x,y) = \left(                                                                                                      \begin{array}{cc}
                                                                                                                    K^{(l)}_{\alpha+\gamma,\alpha+\gamma}(x,x) & K^{(l)}_{\alpha+\gamma,\beta+\gamma}(x,y) \\
                                                                                                                    K^{(l)}_{\beta+\gamma,\alpha+\gamma}(y,x) & K^{(l)}_{\beta+\gamma,\beta+\gamma}(y,y) \\
                                                                                                                  \end{array}
                                                                                                                \right)\,.
\end{equation}
If $\tau$ is the ReLU activation function, \eqref{eq:Kl+1tau} simplifies to
\begin{equation}\label{eq:Kl+1}
K^{(l+1)}_{\alpha,\beta}(x,y) = {\sigma^{(l+1)}_b}^2 + \frac{{\sigma^{(l+1)}_W}^2}{2}\sum_{\gamma\in\mathcal{P}^{(l+1)}}V_{\alpha+\gamma,\beta+\gamma}^{(l)}(x,y)\,,
\end{equation}
where
\begin{equation}\label{eq:Vl}
V^{(l)}_{\alpha,\beta}(x,y) = \sqrt{K^{(l)}_{\alpha,\alpha}(x,x)\,K^{(l)}_{\beta,\beta}(y,y)}\,\Psi\left(\frac{K^{(l)}_{\alpha,\beta}(x,y)}{\sqrt{K^{(l)}_{\alpha,\alpha}(x,x)\,K^{(l)}_{\beta,\beta}(y,y)}}\right)\,,
\end{equation}
with $\Psi:[-1,1]\to\mathbb{R}$ given by
\begin{equation}
\Psi(t) = \frac{\sqrt{1-t^2} + \left(\pi-\arccos t\right)t}{\pi}\,.
\end{equation}
\begin{rem}\label{rem:pos}
Since $\Psi(t)\ge0$ for any $-1\le t\le 1$, we have from \eqref{eq:Vl} and \eqref{eq:Kl+1} that $K^{(l+1)}_{\alpha,\beta}(x,y)\ge0$ for any $x,\,y\in\mathbb{R}^{n^{(0)}}\times\mathcal{I}^{(0)}$ and any $\alpha,\,\beta\in\mathcal{I}^{(l+1)} = \mathcal{I}^{(l)}$.
\end{rem}
Let
\begin{equation}
V^{(l)}(x,y) = \sum_{\alpha\in\mathcal{I}}V^{(l)}_{\alpha,\alpha}(x,y)\,.
\end{equation}
Since $\Psi(1)=1$, we get
\begin{equation}
V^{(l)}(x,x) = K^{(l)}(x,x)\,.
\end{equation}
Since $\Psi(t)\ge t$ for any $-1\le t\le1$, we also get
\begin{equation}
V^{(l)}(x,y) \ge K^{(l)}(x,y)\,.
\end{equation}
Moreover,
\begin{align}
K^{(l+1)}(x,y) &= \left|\mathcal{I}^{(l+1)}\right|{\sigma^{(l+1)}_b}^2 + \frac{\left|\mathcal{P}^{(l+1)}\right|{\sigma^{(l+1)}_W}^2}{2}\,V^{(l)}(x,y)\nonumber\\
&\ge \left|\mathcal{I}^{(l+1)}\right|{\sigma^{(l+1)}_b}^2 + \frac{\left|\mathcal{P}^{(l+1)}\right|{\sigma^{(l+1)}_W}^2}{2}\,K^{(l)}(x,y)\,,
\end{align}
with equality for $y=x$.
We then have
\begin{equation}
K^{(l+1)}(x,x) \ge \frac{\left|\mathcal{P}^{(l+1)}\right|{\sigma^{(l+1)}_W}^2}{2}\,K^{(l)}(x,x) \ge \frac{\left|\mathcal{P}^{(l+1)}\right|{\sigma^{(l+1)}_W}^2\,{C^{(l)}}^2}{2}\left\|x\right\|_2^2\,,
\end{equation}
where we have used the inductive hypothesis.
Moreover,
\begin{equation}
{d^{(l+1)}(x,y)}^2 \le \frac{\left|\mathcal{P}^{(l+1)}\right|{\sigma^{(l+1)}_W}^2}{2}\,{d^{(l)}(x,y)}^2 \le \frac{\left|\mathcal{P}^{(l+1)}\right|{\sigma^{(l+1)}_W}^2\left|\mathcal{I}^{(0)}\right|{C^{(l)}}^2}{2\left|\mathcal{I}^{(l)}\right|}\left\|x-y\right\|_2^2\,,
\end{equation}
where we have used the inductive hypothesis again, and $K^{(l+1)}$ satisfies \eqref{eq:C12} with
\begin{equation}
C^{(l+1)} = {\sigma^{(l+1)}_W}\sqrt{\frac{\left|\mathcal{P}^{(l+1)}\right|}{2}}\,C^{(l)}\,,\qquad M^{(l+1)} = \sqrt{\frac{\left|\mathcal{I}^{(0)}\right|}{\left|\mathcal{I}^{(l)}\right|}} = \sqrt{\frac{\left|\mathcal{I}^{(0)}\right|}{\left|\mathcal{I}^{(l+1)}\right|}}\,.
\end{equation}

\subsection{Skipped Connection}
Let the ($l+1$)-th layer be a skipped connection.
We have
\begin{equation}
K^{(l+1)}_{\alpha,\beta}(x,y) = K^{(l)}_{\alpha,\beta}(x,y) + K^{(l-k)}_{\alpha,\beta}(x,y)\,,
\end{equation}
and
\begin{equation}
K^{(l+1)}(x,y) = K^{(l)}(x,y) + K^{(l-k)}(x,y)\,.
\end{equation}
Since in \autoref{sec:setup} we have imposed $k\le l-2$, we have from \autoref{rem:pos} $K^{(l+1)}_{\alpha,\beta}(x,y)\ge0$ for any $x,\,y\in\mathbb{R}^{n^{(0)}}\times\mathcal{I}^{(0)}$ and any $\alpha,\,\beta\in\mathcal{I}^{(l+1)}$.
We then have
\begin{equation}
K^{(l+1)}(x,x) = K^{(l)}(x,x) + K^{(l-k)}(x,x) \ge \left({C^{(l)}}^2 + {C^{(l-k)}}^2\right)\left\|x\right\|_2^2\,,
\end{equation}
where we have used the inductive hypothesis.
Moreover,
\begin{equation}
{d^{(l+1)}(x,y)}^2 = {d^{(l)}(x,y)}^2 + {d^{(l-k)}(x,y)}^2 \le \frac{\left|\mathcal{I}^{(0)}\right|}{\left|\mathcal{I}^{(l)}\right|}\left({C^{(l)}}^2 + {C^{(l-k)}}^2\right)\left\|x-y\right\|_2^2\,,
\end{equation}
where we have used the inductive hypothesis again, and $K^{(l+1)}$ satisfies \eqref{eq:C12} with
\begin{equation}
C^{(l+1)} = \sqrt{{C^{(l)}}^2 + {C^{(l-k)}}^2}\,,\qquad M^{(l+1)} = \sqrt{\frac{\left|\mathcal{I}^{(0)}\right|}{\left|\mathcal{I}^{(l)}\right|}} = \sqrt{\frac{\left|\mathcal{I}^{(0)}\right|}{\left|\mathcal{I}^{(l+1)}\right|}}\,.
\end{equation}

\subsection{Pooling}
Let the ($l+1$)-th layer be a pooling layer.
Since in the architecture defined in \autoref{sec:setup} we have imposed the $l$-th layer to be a nonlinear convolutional layer, from \autoref{rem:pos} we have $K^{(l)}_{\beta,\gamma}(x,x)\ge0$ for any $\beta,\,\gamma\in\mathcal{I}^{(l)}$.
We can assume from \autoref{thm:master} that $\phi^{(l)}$ is a Gaussian process with covariance $K^{(l)}$.
We then have
\begin{equation}
K^{(l+1)}_{\alpha,\beta}(x,y) = \sum_{\gamma\in\alpha,\,\delta\in\beta}K^{(l)}_{\gamma,\delta}(x,y)\,,\qquad \alpha,\,\beta\in\mathcal{I}^{(l+1)}\,,
\end{equation}
and
\begin{equation}
K^{(l+1)}(x,y) = \sum_{\alpha\in\mathcal{I}^{(l+1)}}\sum_{\beta,\,\gamma\in\alpha}K^{(l)}_{\beta,\gamma}(x,y)\,.
\end{equation}
We have from the inductive hypothesis
\begin{equation}
{d^{(l+1)}(x,y)}^2 \le \frac{\left|\mathcal{I}^{(l)}\right|}{\left|\mathcal{I}^{(l+1)}\right|} {d^{(l)}(x,y)}^2 \le \frac{\left|\mathcal{I}^{(0)}\right|}{\left|\mathcal{I}^{(l+1)}\right|}{C^{(l)}}^2\left\|x-y\right\|_2^2\,.
\end{equation}
Moreover, since $K^{(l)}_{\beta,\gamma}(x,x)\ge0$, we have
\begin{equation}
K^{(l+1)}(x,x) \ge \sum_{\alpha\in\mathcal{I}^{(l+1)}}\sum_{\beta\in\alpha}K^{(l)}_{\beta,\beta}(x,x) = K^{(l)}(x,x) \ge {C^{(l)}}^2\left\|x\right\|_2^2\,,
\end{equation}
where we have used the inductive hypothesis again, and $K^{(l+1)}$ satisfies \eqref{eq:C12} with
\begin{equation}
C^{(l+1)} = C^{(l)}\,,\qquad M^{(l+1)} = \sqrt{\frac{\left|\mathcal{I}^{(0)}\right|}{\left|\mathcal{I}^{(l+1)}\right|}}\,.
\end{equation}
\subsection{Flattening}
From \autoref{thm:master}, we can assume that $\phi^{(L)}$ is the centered Gaussian process with covariance $K^{(L)}$.
The proof is completely analog to the proof in \autoref{sec:conv}, and $K^{(L+1)}$ satisfies \eqref{eq:C12} with
\begin{equation}
C = C^{(L+1)} = {\sigma^{(L+1)}_W}\sqrt{\frac{\left|\mathcal{I}^{(L_f)}\right|}{2}}\,C^{(L)}\,,\qquad M = M^{(L+1)} = \sqrt{\frac{\left|\mathcal{I}^{(0)}\right|}{\left|\mathcal{I}^{(L_f)}\right|}}\,.
\end{equation}

\section{Proof of \autoref{thm:Nepsilon}}\label{sec:thmNepsilonproof}
The upper bound for $0<\epsilon\le\frac{1}{\sqrt{n}}$ has been proven in \cite{price2016sublinear}.
Let $\frac{1}{\sqrt{n}}<\epsilon<1$, and let $m\in\mathbb{N}$ be such that
\begin{equation}
2\le m \le n\,,\qquad \frac{1}{m}\le \epsilon^2<\frac{1}{m-1}\,.
\end{equation}
We consider the lattice
\begin{equation}
\mathcal{L}_m = \frac{\mathbb{Z}^n}{m}\cap\mathcal{B}_1\,.
\end{equation}
For any $x\in\mathcal{B}_1$, there exists $y\in\mathcal{L}_m$ such that for any $i=1,\,\ldots,\,n$
\begin{equation}
\left|x_i-y_i\right|\le\min\left(\frac{1}{m},\,|x_i|\right)\,.
\end{equation}
We have
\begin{align}
\left(x-y\right)^2 &\le \frac{1}{m^2}\,\left|\left\{i:|x_i|>\frac{1}{m}\right\}\right| + \sum_{i\,:\,|x_i|\le\frac{1}{m}}|x_i|^2 \le \frac{1}{m}\sum_{i:|x_i|>\frac{1}{m}}|x_i| + \frac{1}{m}\sum_{i\,:\,|x_i|\le\frac{1}{m}}|x_i| = \frac{\|x\|_1}{m}\nonumber\\
&\le \frac{1}{m} \le \epsilon^2\,,
\end{align}
therefore $N(\epsilon)\le|\mathcal{L}_m|$.
The claim follows since
\begin{equation}
|\mathcal{L}_m| = \sum_{k=0}^{m-1}2^k\binom{n}{k}\binom{m-1}{k}\le \left(2\,n\right)^{m-1} \le \left(2\,n\right)^\frac{1}{\epsilon^2}\,.
\end{equation}

\section{Lemmas}
\subsection{Proof of \autoref{lem:dh}}\label{sec:lemdhproof}
Since $\hat{K}\le K$ we have
\begin{align}
{d(x,y)}^2 &= \left(
             \begin{array}{c}
               1 \\
               -1 \\
             \end{array}
           \right)\left(
                    \begin{array}{cc}
                      K(x,x) & K(x,y) \\
                      K(y,x) & K(y,y) \\
                    \end{array}
                  \right)\left(
                           \begin{array}{cc}
                             1 & -1 \\
                           \end{array}
                         \right)\nonumber\\
                         &\ge \left(
             \begin{array}{c}
               1 \\
               -1 \\
             \end{array}
           \right)\left(
                    \begin{array}{cc}
                      \hat{K}(x,x) & \hat{K}(x,y) \\
                      \hat{K}(y,x) & \hat{K}(y,y) \\
                    \end{array}
                  \right)\left(
                           \begin{array}{cc}
                             1 & -1 \\
                           \end{array}
                         \right) = {\hat{d}(x,y)}^2\,.
\end{align}

\subsection{Proof of \autoref{lem:Kr}}\label{sec:lemKrproof}
We have for any $x\in \mathcal{B}_r^1$
\begin{align}
\left|K(x,x_0) - K(x_0,x_0)\right| &= \left|\left(\Phi(x)-\Phi(x_0)\right)\cdot\Phi(x_0)\right| \le \left\|\Phi(x)-\Phi(x_0)\right\|\left\|\Phi(x_0)\right\|\nonumber\\
&\le M\,C\left\|x-x_0\right\|_2\sqrt{K(x_0,x_0)} \le M\,C\,r\sqrt{K(x_0,x_0)}\,,
\end{align}
where we have used \eqref{eq:C12} and that $\left\|x-x_0\right\|_2\le r$, and the claim follows.

\subsection{Proof of \autoref{lem:sigmar}}\label{sec:lemsigmarproof}
We have for any $x\in \mathcal{B}_r^1$
\begin{equation}
\hat{K}(x,x) = \left\|\Phi(x)-\Phi(x_0)\right\|^2 - \frac{\left(K(x_0,x_0)-K(x,x_0)\right)^2}{K(x_0,x_0)} \le M^2\,C^2\left\|x-x_0\right\|^2 \le M^2\,C^2\,r^2\,,
\end{equation}
where we have used \eqref{eq:C12}.

\section{Experimental Details}
\label{app:experimental_details}

\subsection{Adversarial Attack Methods}
\label{app:attack_methods}

To find adversarial examples, various different adversarial attack methods were used. The list of adversarial attack used for each norm are given in \autoref{tab:adv_attack_list}. Hyperparameters for adversarial attack algorithms were also varied to find adversarial examples at the smallest norm possible. All attacks were performed using the python package Foolbox \cite{rauber_foolbox_2018}.

For adversarial attacks on random neural networks (\autoref{sec:adv_attack_random}), attacks were performed on random inputs where each input was sampled from the uniform distribution bounded by $[0,1]$. For random neural networks, weights were chosen randomly according to a normal distribution with variance scaled depending on the number of neurons in prior and posterior layers \cite{glorot2010understanding}. For each random neural network constructed, attacks were performed on batches of 10 randomized inputs. To be consistent with attacks in the adversarial literature, neural networks were constructed with two output neurons to perform binary classification. No activation (\textit{e.g.,} softmax) was included in the final layer and attacks were directly performed on output logits.

Inputs were bounded by $[0,1]$ in every dimension or pixel. Thus, attack algorithms were restricted in their operation within these bounds. In the case of random inputs, inputs were adversarially attacked to change their binary classification. In the cases where train or test data was provided (\textit{i.e.,} in the cases of MNIST and CIFAR10), inputs were adversarially attacked to change the classification provided by the network (not necessarily the correct classification).

\begin{table}[ht]
\begin{tabular}{ll}
\hline
\multicolumn{2}{l}{\textbf{Adversarial Attack Methods}}         \\ \hline
\textbf{$\ell _1$ Norm Attacks}   & EAD Attack \cite{zhao_admm-based_2018}                    \\
                             & Pointwise Attack \cite{schott_towards_2018}                \\
                             & Saliency Map Attack \cite{papernot_limitations_2015}             \\
                             & Sparse L1 Basic Iterative Method \cite{tramer_adversarial_2019} \\ \hline
\textbf{$\ell _2$ Norm Attacks}   & Basic Iterative Method \cite{kurakin2018adversarial}          \\
                             & Carlini Wagner Attack \cite{carlini_towards_2017}           \\
                             & Decoupled Direction Norm Attack \cite{rony_decoupling_2019} \\ \hline
\textbf{$\ell _\infty$ Norm Attacks} & Basic Iterative Method \cite{kurakin2018adversarial}          \\
                             & Momentum Iterative Method \cite{dong_boosting_2018}       \\
                             & Adam Projected Gradient Descent \cite{madry_towards_2019,carlini_towards_2017} \\ \hline
\end{tabular}
\caption{List of attack algorithms used to find the closest adversarial example for each norm. Attack algorithms were performed with varying hyperparameters. Among the adversarial examples given by the various attacks, the adversarial example with the smallest distance norm from the starting point is assumed to be the closest adversarial example.}
\label{tab:adv_attack_list}
\end{table}

\subsection{Network Architectures}
\label{app:net_architectures}

Various different networks were empirically studied. Layer sequences for the various networks are provided in \autoref{tab:net_architectures}. Weights in all networks were initialized randomly with variance inversely proportional to the size of the previous layer, often termed He initialization \cite{he_delving_2015}. In all cases, to conform to the standards provided in the Foolbox toolbox \cite{rauber_foolbox_2018}, output layers contained two neurons, one for each class in the binary classification task. In all cases, the nonlinear activation used was the rectified linear unit (ReLU).

Inputs to all the networks are assumed to be 2-dimensional images with 3 channels. The only exception to this case is for networks trained on MNIST data where inputs have only one channel.

\begin{table}[ht]
\small
\centering
\setlength{\extrarowheight}{0pt}
\addtolength{\extrarowheight}{\aboverulesep}
\addtolength{\extrarowheight}{\belowrulesep}
\setlength{\aboverulesep}{0pt}
\setlength{\belowrulesep}{0pt}
\arrayrulecolor[rgb]{0.753,0.753,0.753}
\begin{tabular}{|c|c|c|c|}
\toprule
\rowcolor[rgb]{0.835,0.835,0.835} \multicolumn{1}{!{\color{black}\vrule}c!{\color{black}\vrule}}{Simplified Residual Network} & \multicolumn{1}{c!{\color{black}\vrule}}{\begin{tabular}[c]{@{}>{\cellcolor[rgb]{0.835,0.835,0.835}}c@{}}``LeNet'' Style \cite{lecun_gradient-based_1998} \\Network\end{tabular}} & \multicolumn{1}{c!{\color{black}\vrule}}{\begin{tabular}[c]{@{}>{\cellcolor[rgb]{0.835,0.835,0.835}}c@{}} Fully Connected \\Network\end{tabular}}               & \multicolumn{1}{c!{\color{black}\vrule}}{\begin{tabular}[c]{@{}>{\cellcolor[rgb]{0.835,0.835,0.835}}c@{}} Simple Convolutional \\Network\end{tabular}}    \\
\arrayrulecolor{black}\hline
\begin{tabular}[c]{@{}c@{}}Residual Block*\\ (32 channels)\end{tabular}                                                   & \begin{tabular}[c]{@{}c@{}}3x3 Conv - ReLU\\(128 Channels) \end{tabular}                                                                                               & Flatten                                                                         & \begin{tabular}[c]{@{}c@{}}3x3 Conv - ReLU\\(128 Channels)\end{tabular}  \\
\arrayrulecolor[rgb]{0.753,0.753,0.753}\hline
\begin{tabular}[c]{@{}c@{}}Residual Block**\\(64 channels)\end{tabular}                                                   & \begin{tabular}[c]{@{}c@{}}3x3 Conv - ReLU\\(128 Channels)\end{tabular}                                                                                                & \begin{tabular}[c]{@{}c@{}}Fully Connected - ReLU\\(100 Neurons)\end{tabular}   & \begin{tabular}[c]{@{}c@{}}3x3 Conv - ReLU\\(128 Channels)\end{tabular}  \\
\hline
\begin{tabular}[c]{@{}c@{}}Residual Block**\\(128 channels)\end{tabular}                                                  & \begin{tabular}[c]{@{}c@{}}2x2 Average Pooling\end{tabular}                                                                                               & \begin{tabular}[c]{@{}c@{}}Fully Connected - ReLU \\(100 Neurons) \end{tabular} & \begin{tabular}[c]{@{}c@{}}3x3 Conv - ReLU\\(128 Channels)\end{tabular}  \\
\hline
Flatten                                                                                                                   & \begin{tabular}[c]{@{}c@{}}3x3 Conv - ReLU\\(128 Channels)\end{tabular}                                                                                               & Output Layer                                                                    & \begin{tabular}[c]{@{}c@{}}3x3 Conv - ReLU\\(128 Channels)\end{tabular}  \\
\hline
Output Layer                                                                                                              & \begin{tabular}[c]{@{}c@{}}3x3 Conv - ReLU\\(128 Channels)\end{tabular}                                                                                                                          &                                                                                 & Flatten                                                                  \\
\hline
                                                                                                                          & 2x2 Average Pooling                                                                                                                                                         & \multicolumn{1}{l|}{}                                                           & Output Layer                                                             \\
\hline  & Flatten   &   &   \\
\hline  & \begin{tabular}[c]{@{}c@{}}Fully Connected - ReLU\\(512 Neurons)\end{tabular}    &   &   \\
\hline  & \begin{tabular}[c]{@{}c@{}}Fully Connected - ReLU\\(512 Neurons)\end{tabular}    &   &   \\
\hline  & Output Layer    &   &   \\
\hline

\multicolumn{4}{l}{\begin{tabular}[c]{@{}l@{}} * residual blocks contain 2 3x3 convolutional layers each followed by ReLU activation\\ ** first convolution layer in residual block has stride set to 2 (feature map size is halved)\end{tabular}}                                                                                                                                                                                                              \\
\end{tabular}
\arrayrulecolor{black}
\caption{Layer sequences for the various networks empirically studied. Layer sequences should be read from top to bottom.}
\label{tab:net_architectures}
\end{table}

\subsection{Trained Networks}
\label{app:trained_networks}

For analysis on trained networks (\autoref{sec:adv_attack_trained}), networks were trained on either MNIST or CIFAR10 data under the task of binary classification. A softmax activation was placed on the two neurons of the last layer of all networks. For the case of MNIST, the binary classification task was determining if a digit is odd or even. For CIFAR10, image classes were assigned to binary categories of either \{airplane, bird, deer, frog, ship\} or \{automobile, cat, dog, horse, truck\}.  All networks were trained to minimize categorical cross-entropy using the Adam optimizer \cite{kingma2014adam}. Batch normalization was not included in any networks and none of the networks used dropout \cite{srivastava2014dropout}. For CIFAR10 data, networks were trained for 25 epochs on the complete training set with a learning rate of 0.0001. For MNIST data, networks were trained for 15 epochs on the complete training set with a learning rate of 0.0001. For the simplified residual network, trained networks achieved an average binary classification accuracy of 99.8\% and 99.4\% on the MNIST training and test sets respectively. Furthermore, the same network architecture trained on CIFAR10 achieved an average of 98.8\% and 82.5\% accuracy on training and test sets.

Adversarial attacks were performed on batches of 20 random images or 20 randomly chosen images from the training or test sets. Median adversarial distances in \autoref{fig:trained_resnet_profile} were taken from a sample size of 5000 points -- 250 trained networks each attacking 20 random images.


\section{Supplementary Figures}
\label{app:supplementary_figures}

Adversarial attacks on random networks were performed on various architectures. In this section, we include figures for attacks performed on networks not included in \autoref{sec:adv_attack_random}.

\begin{figure}[ht]
\centering
  \includegraphics[]{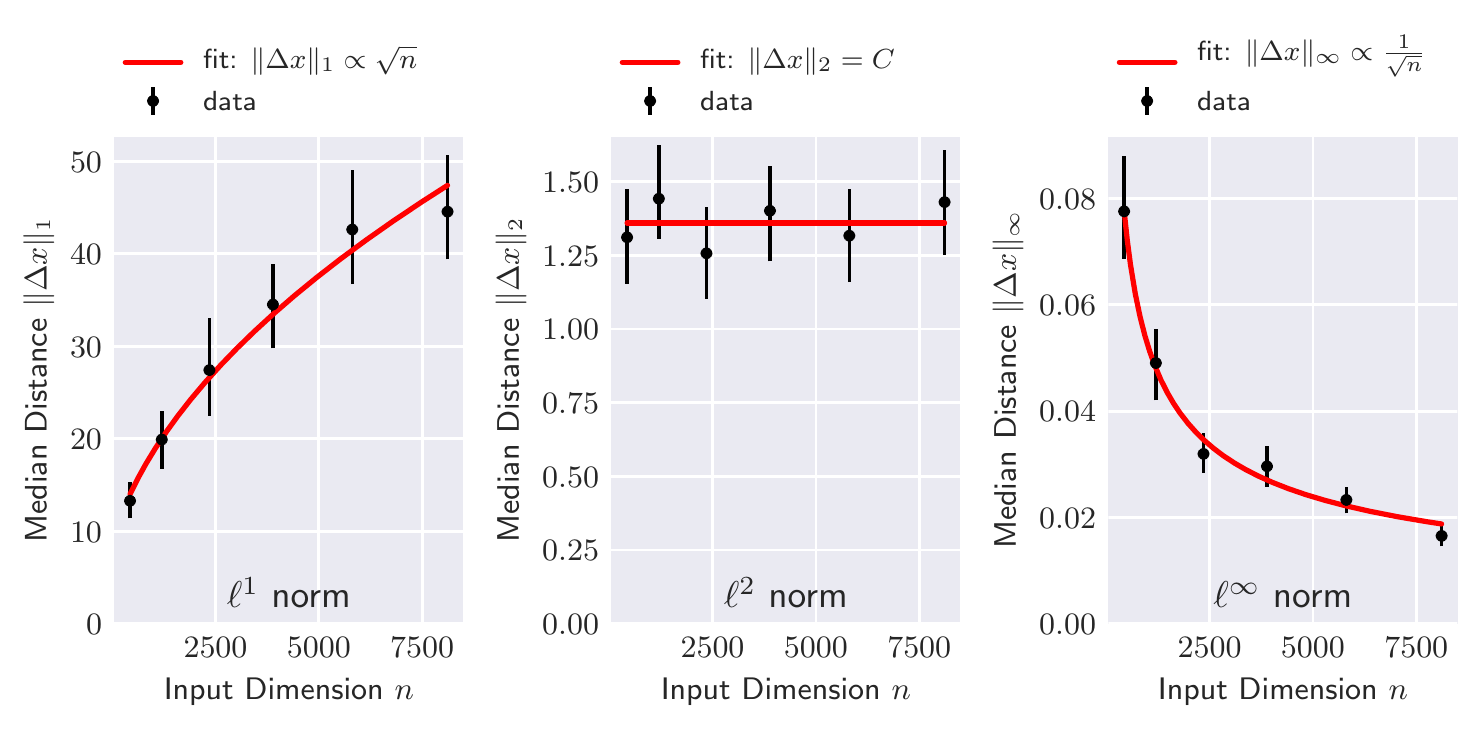}
  \label{fig:random_LeNet}
  \caption{{\bf Random untrained convolutional networks:} The median $\ell^p$ distances of closest adversarial examples from their respective inputs for $p =1,\,2,\,\infty$ scale as predicted in \autoref{rem:norm_p_scaling} for a ``LeNet'' \cite{lecun_gradient-based_1998} architecture (see \autoref{app:net_architectures} for full description of network). Error bars span $\pm 5$ percentiles from the median. For each input dimension, results are calculated from 2000 samples (200 random networks each attacked at 10 random points). See \autoref{app:experimental_details} for further details on how experiments were performed.}
\end{figure}

\begin{figure}[ht]
\centering
  \includegraphics[]{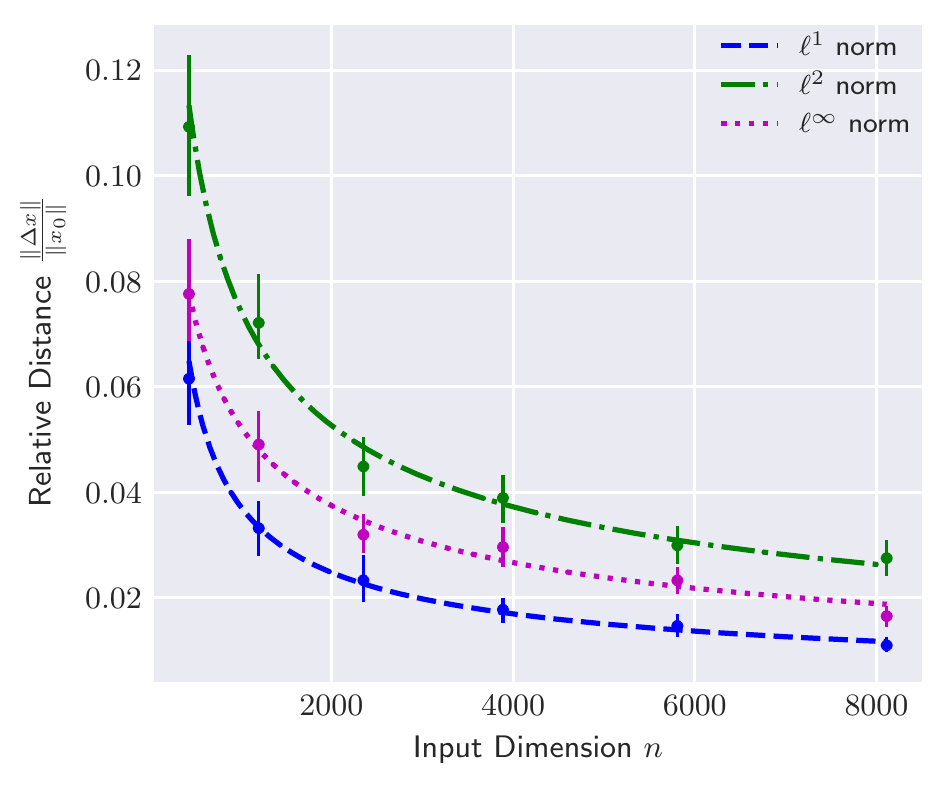}
  \label{fig:random_LeNet_normalized}
  \caption{{\bf Random untrained networks:} Median relative distance of closest adversarial examples $\| \Delta x \|_p / \| x_0 \|_p$ from their respective inputs ($p \in \{1,\,2,\,\infty\}$) scale with the input dimension $n$ as $O(1/\sqrt{n})$ in all norms for a ``LeNet'' \cite{lecun_gradient-based_1998} architecture (see \autoref{app:net_architectures} for full description of network). Error bars span $\pm 5$ percentiles from the median. For each input dimension, results are calculated from 2000 samples (200 random networks each attacked at 10 random points).}
\end{figure}

\begin{figure}[ht]
\centering
  \includegraphics[]{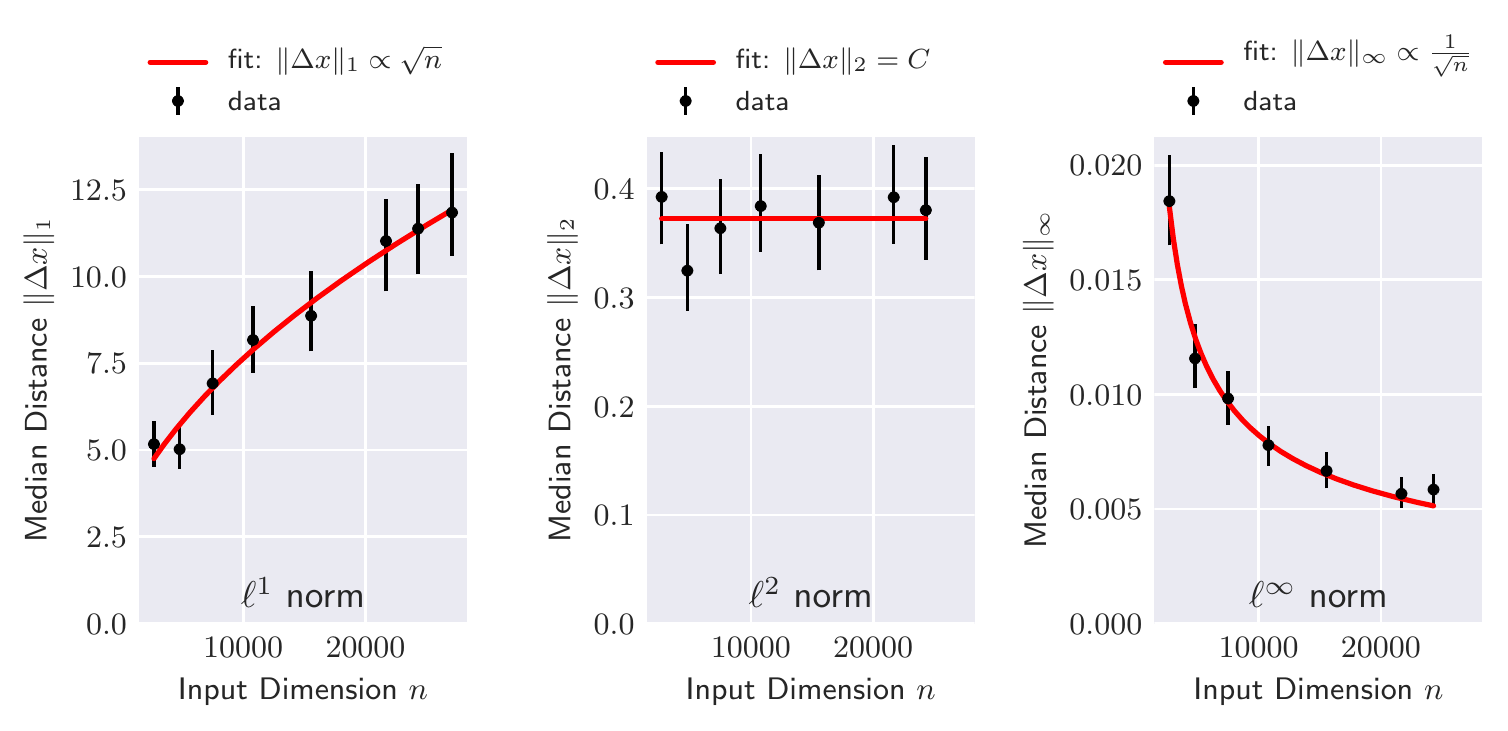}
  \label{fig:random_FCN}
  \caption{{\bf Random untrained networks:} Median distance of closest adversarial examples $\| \Delta x \|_p$ from their respective inputs ($p \in \{1,\,2,\,\infty\}$) scale as predicted in \autoref{rem:norm_p_scaling} for a fully connected network (see \autoref{app:net_architectures} for full description of network). Error bars span $\pm 5$ percentiles from the median. For each input dimension, results are calculated from 2000 samples (200 random networks each attacked at 10 random points). See \autoref{app:experimental_details} for further details on how experiments were performed.}
\end{figure}

\begin{figure}[ht]
\centering
  \includegraphics[]{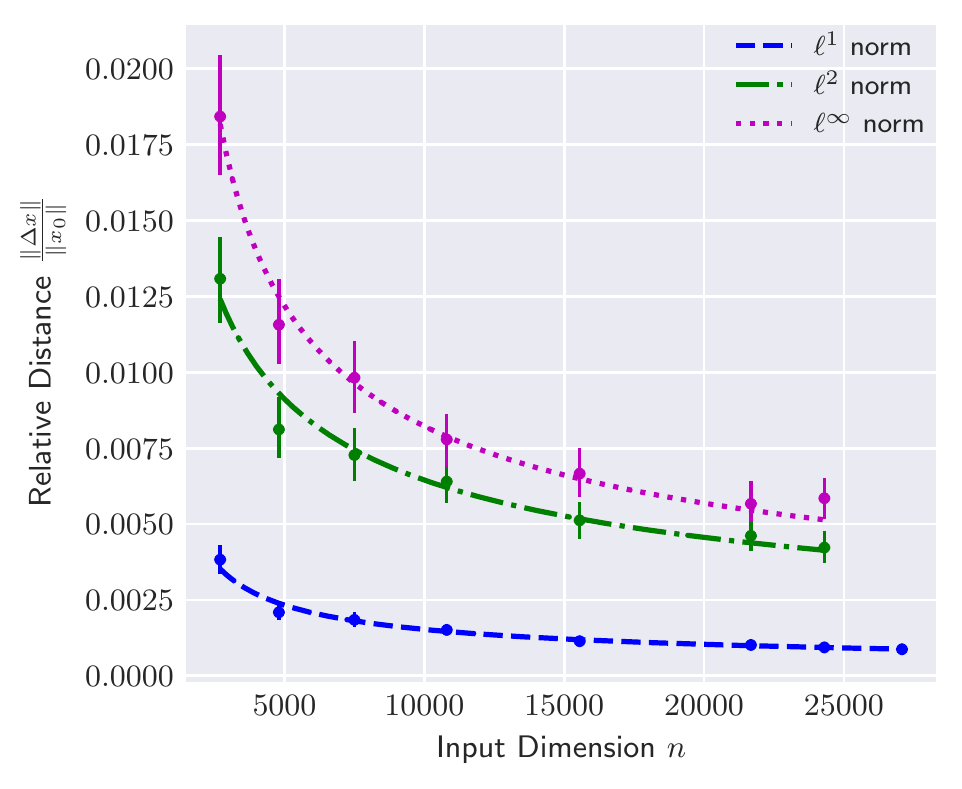}
  \label{fig:random_FCN_normalized}
  \caption{{\bf Random untrained networks:} Median relative distance of closest adversarial examples $\| \Delta x \|_p / \| x_0 \|_p$ from their respective inputs ($p \in \{1,\,2,\,\infty\}$) scale with the input dimension $n$ as $O(1/\sqrt{n})$ in all norms for a fully connected network (see \autoref{app:net_architectures} for full description of network). Error bars span $\pm 5$ percentiles from the median. For each input dimension, results are calculated from 2000 samples (200 random networks each attacked at 10 random points).}
\end{figure}

\begin{figure}[ht]
\centering
  \includegraphics[]{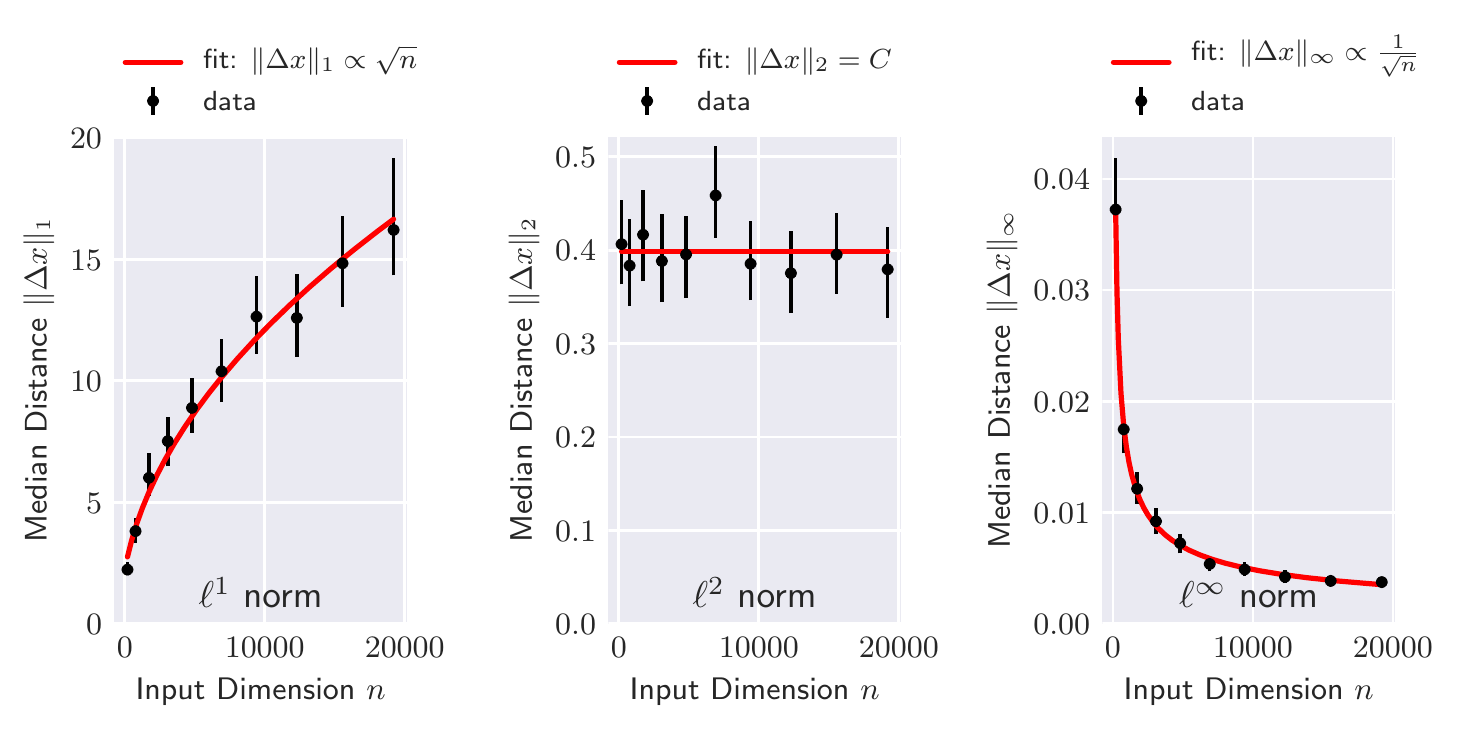}
  \label{fig:random_CNN}
  \caption{{\bf Random untrained networks:} Median distance of closest adversarial examples $\| \Delta x \|_p$ from their respective inputs ($p \in \{1,\,2,\,\infty\}$) scale as predicted in \autoref{rem:norm_p_scaling} for a simple convolutional network (see \autoref{app:net_architectures} for full description of network). Error bars span $\pm 5$ percentiles from the median. For each input dimension, results are calculated from 2000 samples (200 random networks each attacked at 10 random points). See \autoref{app:experimental_details} for further details on how experiments were performed. }
\end{figure}

\begin{figure}[ht]
\centering
  \includegraphics[]{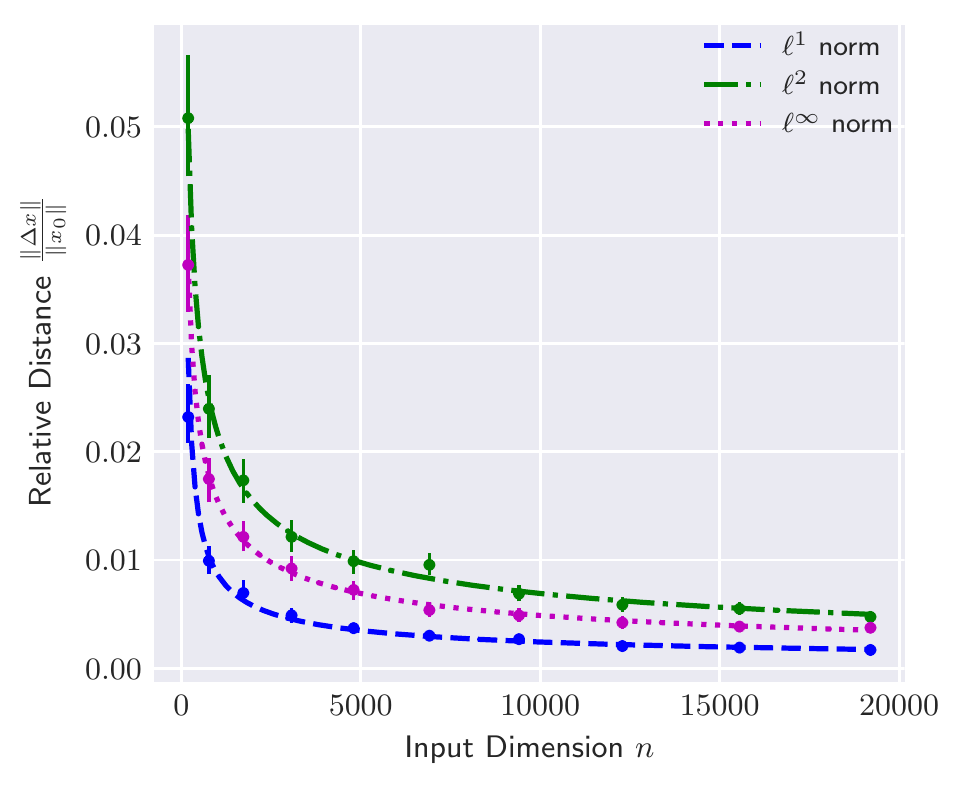}
  \label{fig:random_CNN_normalized}
  \caption{{\bf Random untrained networks:} Median relative distance of closest adversarial examples $\| \Delta x \|_p / \| x_0 \|_p$ from their respective inputs ($p \in \{1,\,2,\,\infty\}$) scale with the input dimension $n$ as $O(1/\sqrt{n})$ in all norms for a simple convolutional network (see \autoref{app:net_architectures} for full description of network). Error bars span $\pm 5$ percentiles from the median. For each input dimension, results are calculated from 2000 samples (200 random networks each attacked at 10 random points).}
\end{figure}

\clearpage

\bibliography{biblio}
\bibliographystyle{icml2021}

\end{document}